\documentclass[dvipsnames]{article}
\usepackage{colm2024_conference}
\usepackage[utf8]{inputenc}
\usepackage[T1]{fontenc}
\usepackage{microtype}
\usepackage{amsmath,amssymb,mathtools,amsthm}
\usepackage{booktabs,multirow,makecell,tabularx}
\usepackage{longtable}
\usepackage{graphicx}
\usepackage{float}
\usepackage{caption}
\usepackage{subcaption}
\usepackage{enumitem}
\usepackage{listings}
\usepackage[most]{tcolorbox}
\usepackage{pifont}
\usepackage{xspace}
\usepackage{xurl}
\usepackage{siunitx}
\usepackage{adjustbox}
\usepackage[section]{placeins}


\usepackage{amsmath,amsfonts,bm}

\def\eqref#1{equation~\ref{#1}}

\def\1{\bm{1}}

\DeclareMathAlphabet{\mathsfit}{\encodingdefault}{\sfdefault}{m}{sl}
\SetMathAlphabet{\mathsfit}{bold}{\encodingdefault}{\sfdefault}{bx}{n}

\usetikzlibrary{arrows.meta,positioning}

\providecommand{\bloglink}{}
\renewcommand{\bloglink}{https://unipat.ai/blog/ScienceVision}
\newcommand{\codelink}{https://github.com/UniPat-AI/ScienceVision}

\newcommand{\code}[1]{\begingroup\urlstyle{tt}\path{#1}\endgroup}

\definecolor{taskbackground}{RGB}{247,248,250}
\definecolor{taskframe}{RGB}{75,85,99}
\newtcolorbox{taskbox}[1]{
  enhanced,
  breakable,
  colback=taskbackground,
  colframe=taskframe,
  boxrule=0.5pt,
  arc=1mm,
  left=5pt,
  right=5pt,
  top=4pt,
  bottom=5pt,
  title={#1},
  fonttitle=\bfseries,
  coltitle=white,
  colbacktitle=taskframe
}

\lstdefinestyle{sciencevisioncode}{
  basicstyle=\ttfamily\footnotesize,
  breaklines=true,
  columns=fullflexible,
  keepspaces=true,
  showstringspaces=false,
  frame=single,
  rulecolor=\color{taskframe},
  backgroundcolor=\color{taskbackground}
}

\newcommand{\bench}{\textit{ScienceVision}}
\newcommand{\benchfirst}{\textbf{ScienceVision}}

\newif\ificlrfinal
\iclrfinaltrue

\ExplSyntaxOn
\NewDocumentCommand{\ScienceVisionTitleTracking}{m}{
  \group_begin:
  \seq_set_split:Nnn \l_tmpa_seq {~}{#1}
  \seq_map_indexed_inline:Nn \l_tmpa_seq {
    \int_compare:nNnT {##1}>{1}{\space\kern -.4pt}
    \str_map_inline:nn {##2}{####1\kern -.4pt}
  }
  \unkern
  \group_end:
}
\ExplSyntaxOff
\title{\texorpdfstring{\ScienceVisionTitleTracking{ScienceVision: Can AI Observe and Reason Like a Scientist?}}{ScienceVision: Can AI Observe and Reason Like a Scientist?}}
\author{UniPat AI, Elorian AI}

\begin{document}
\maketitle

\begin{abstract}
Scientific discovery builds on accurate observations interpreted through domain knowledge and reasoning. For scientific 3D data, this requires understanding structures and spatial relationships, from protein contacts to anatomical organization. Existing multimodal benchmarks largely evaluate text and 2D images, while scientific 3D evaluations vary in representations and protocols across disciplines, limiting cross-disciplinary comparison.
To assess these foundational capabilities, we introduce \benchfirst{}, a cross-domain benchmark for scientific 3D understanding and reasoning through expert-level visual question answering (VQA). We construct 250 questions, with 50 each from Biology, Chemistry, Medical Imaging, Mechanical CAD, and Architecture, through a pipeline combining synthetic question generation with domain-expert review and annotation. 
Each task pairs a native 3D artifact with a scientific question, a specified response format, and a task-specific reference answer, requiring domain knowledge to be applied to geometric or topological evidence.
We evaluate 12 models on the same questions in three settings: visual question answering with native 3D artifacts accessed through an agent harness's file and code-execution tools; the same workflow augmented with expert-inspired viewer inspection and measurement tools exposed through the Model Context Protocol (MCP); and visual question answering from prescribed rendered 2D images using an agent harness. 
All settings use the same task-specific deterministic final-answer scoring. The inspection environment supports view changes, target isolation, and slice inspection.
In this main comparison, the best mean answer score is $0.53$ out of $1$. Mean scores in the native-artifact workflow without additional MCP tools exceed those in the rendered-image workflow for every model, by $17.2$ percentage points on average.
Our code and benchmark data are available at {\hypersetup{urlcolor=citegray}\href{\codelink}{\raisebox{-0.15em}{\includegraphics[height=1em]{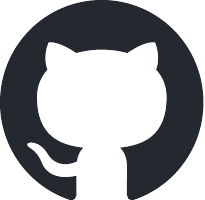}}\,UniPat-AI/ScienceVision}} to support reproduction of our results.

\end{abstract}

\begin{figure}[H]
\centering
\includegraphics[width=\linewidth]{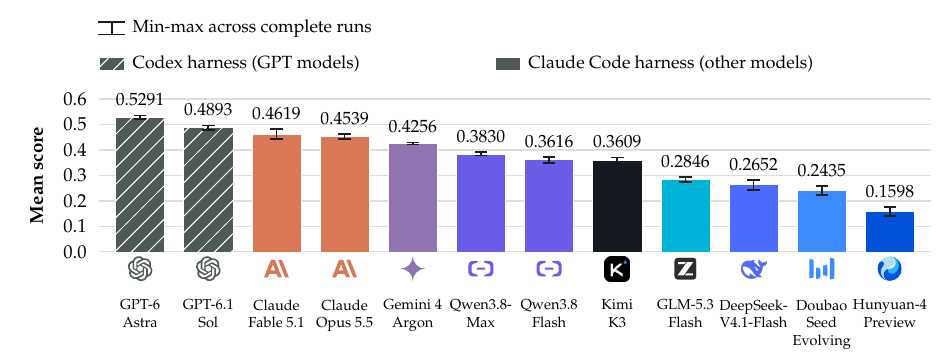}
\caption{\textbf{ScienceVision performance on scientific questions about native 3D artifacts.} Each bar represents one of 12 model--harness configurations answering 250 questions with file access and code execution. Bar height is the mean task score from 0 to 1 including partial credit; higher is better. Whiskers show the min--max of complete-run means over the same 250 tasks. Bars descend by unrounded mean; labels show four decimals. Hatched GPT bars use Codex and solid bars use Claude Code.}
\label{newfig:leaderboard}
\end{figure}

\section{Introduction}
\label{newsec:intro}

Scientific discovery depends on finding evidence that bears on a question and interpreting it using domain knowledge. For questions about three-dimensional scientific objects, relevant evidence may be hidden by occlusion or distributed across views and slices. Researchers may need to examine the object from another viewpoint or inspect a different slice before making a structural judgment.
For an AI system answering such questions, preselected images fix which views or slices it can examine. An evaluation using only those images cannot assess how the model would select and obtain additional observations beyond that set. Molecular information can be distributed across views \citep{Xiang2024VideoMol}. A molecular question, for example, may require changing the view to reveal relevant atoms before analyzing a bonded path or measuring an angle. Supplying the native object with inspection or analysis tools lets models seek further evidence, subject to the operations available in each setting.

Scientific visual question answering (VQA) provides a way to study how models connect structural evidence to domain-specific answers \citep{Burgess2025MicroVQA,Mandiga2026ProtStructQA}. The questions define the requested scientific judgments, and the access conditions specify how models can examine the supporting objects. Applying common answer criteria to the same questions allows final-answer performance to be compared across these conditions.
DeepTumorVQA evaluates abdominal CT questions \citep{Chen2026DeepTumorVQA}, and ProtStructQA evaluates protein structural questions \citep{Mandiga2026ProtStructQA}; both are limited to a single scientific domain. SciVisAgentBench spans multiple domains, but combines analysis and visualization goals and uses both multimodal judges and deterministic evaluators \citep{Ai2026SciVisAgentBench}. These designs leave a gap in cross-domain scientific 3D question answering: a focused comparison of evidence-access settings with deterministic answer checks.

We introduce \benchfirst{}, a benchmark for scientific 3D understanding and reasoning formulated as VQA. Its questions cover Architecture, Biology, Chemistry, Mechanical CAD, and Medical Imaging. Each task pairs a native 3D object with a question, explicit response requirements, and a reference answer. Submitted answers are scored deterministically using predefined, task-specific matching rules, including accepted equivalents and numerical tolerances where specified. Expert-level describes the domain knowledge and conventions required to interpret the tasks.
Models answer the VQA questions using agent harnesses that manage file access, code execution, and tool interactions. In \textbf{Harness + 3D}, models work with native 3D artifacts through the harness's file and code-execution tools. \textbf{Harness + 3D + MCP} additionally exposes viewer-specific tools for changing views, isolating targets, navigating slices, and measuring geometry \citep{ModelContextProtocol2025Spec}. In \textbf{Harness + 2D}, models answer the questions from rendered 2D images using an agent harness. All settings use the same VQA questions, response requirements, and deterministic final-answer scoring rules.

We compare 12 models under these conditions. Harness + 3D yields higher scores than Harness + 2D for every model. The score changes associated with adding MCP tools are smaller than the Harness + 3D--Harness + 2D differences for every model and vary in direction. These comparisons concern the evaluated model--harness configurations, which differ in available evidence and permitted computation. Domain and task-family analyses, reference sensitivity checks, and supplementary execution settings further qualify the results.

Our contributions are:
\begin{itemize}[leftmargin=*]
    \item \textbf{A cross-disciplinary scientific 3D VQA benchmark.} \bench{} provides 250 expert-level questions across five equally represented domains and 22 subdomains. Each task pairs a native 3D object with a domain-specific question, explicit response requirements, and a structured reference answer. Some tasks connect target identification under occlusion to geometric or topological reasoning.
    \item \textbf{An expert-inspired MCP inspection environment.} Viewer-dependent interfaces expose viewpoint control, target isolation, slice browsing, and scientific measurement. Designed around expert inspection workflows, these interfaces let models select and inspect task-relevant evidence during question answering.
    \item \textbf{A unified protocol for deterministic scientific VQA evaluation.} Task-specific rules verify scored answer components against reference answers and assign partial credit where specified without a model judge. The same questions, response requirements, and scoring rules apply to Harness + 3D, Harness + 3D + MCP, and Harness + 2D. Results are reported by evaluation setting and domain.
\end{itemize}

\providecommand{\sciencevisionDatasetAppendix}{newapp:tasks}
\section{The ScienceVision Benchmark}
\label{newsec:tasks}
\label{sec:benchmark}

\begin{figure}[!htbp]
\centering
\begin{minipage}[t]{0.19\linewidth}
\centering
\parbox[c][33mm][c]{\linewidth}{\centering
\includegraphics[width=\linewidth,height=33mm,keepaspectratio,trim=0bp 29bp 124bp 280bp,clip]{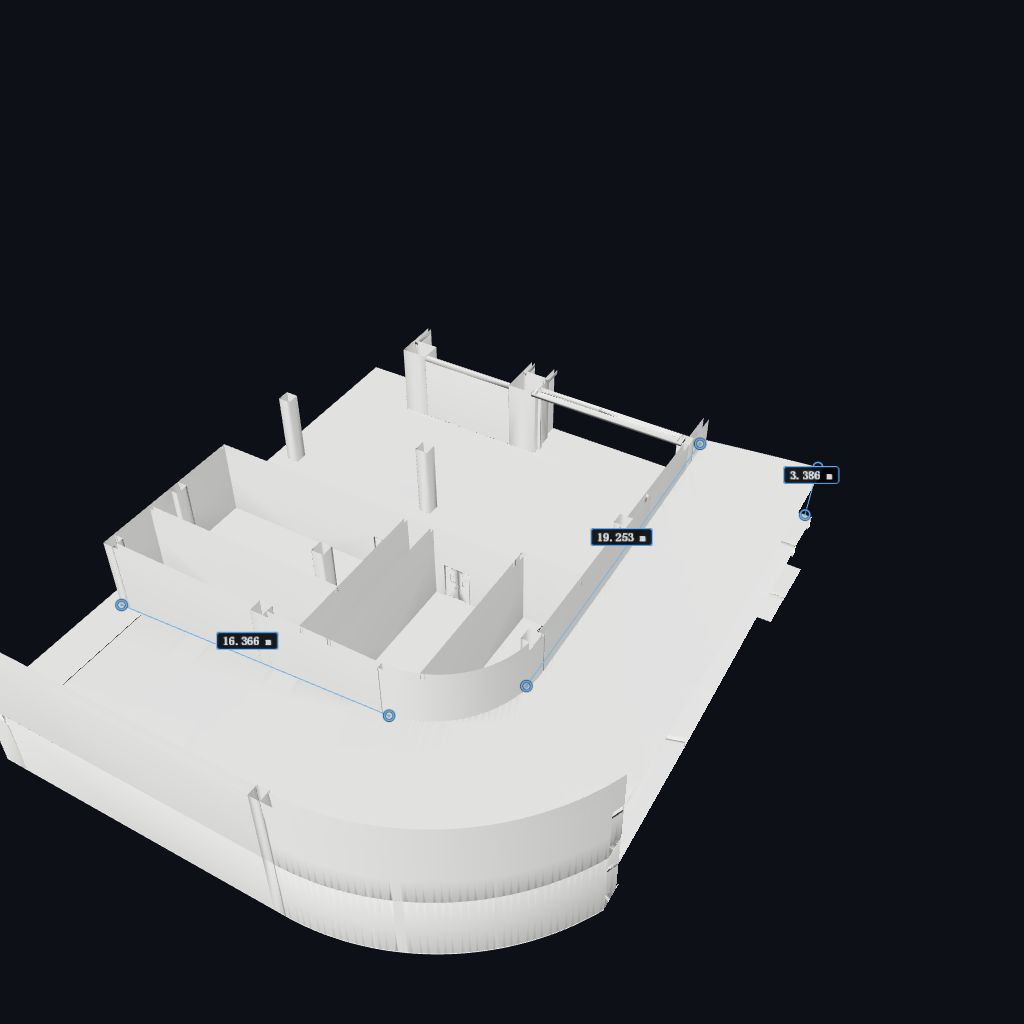}
}
\par\textbf{Architecture}
\par Cutaway
\end{minipage}\hfill
\begin{minipage}[t]{0.19\linewidth}
\centering
\parbox[c][33mm][c]{\linewidth}{\centering
\includegraphics[width=\linewidth,height=33mm,keepaspectratio,trim=10bp 173bp 158bp 130bp,clip]{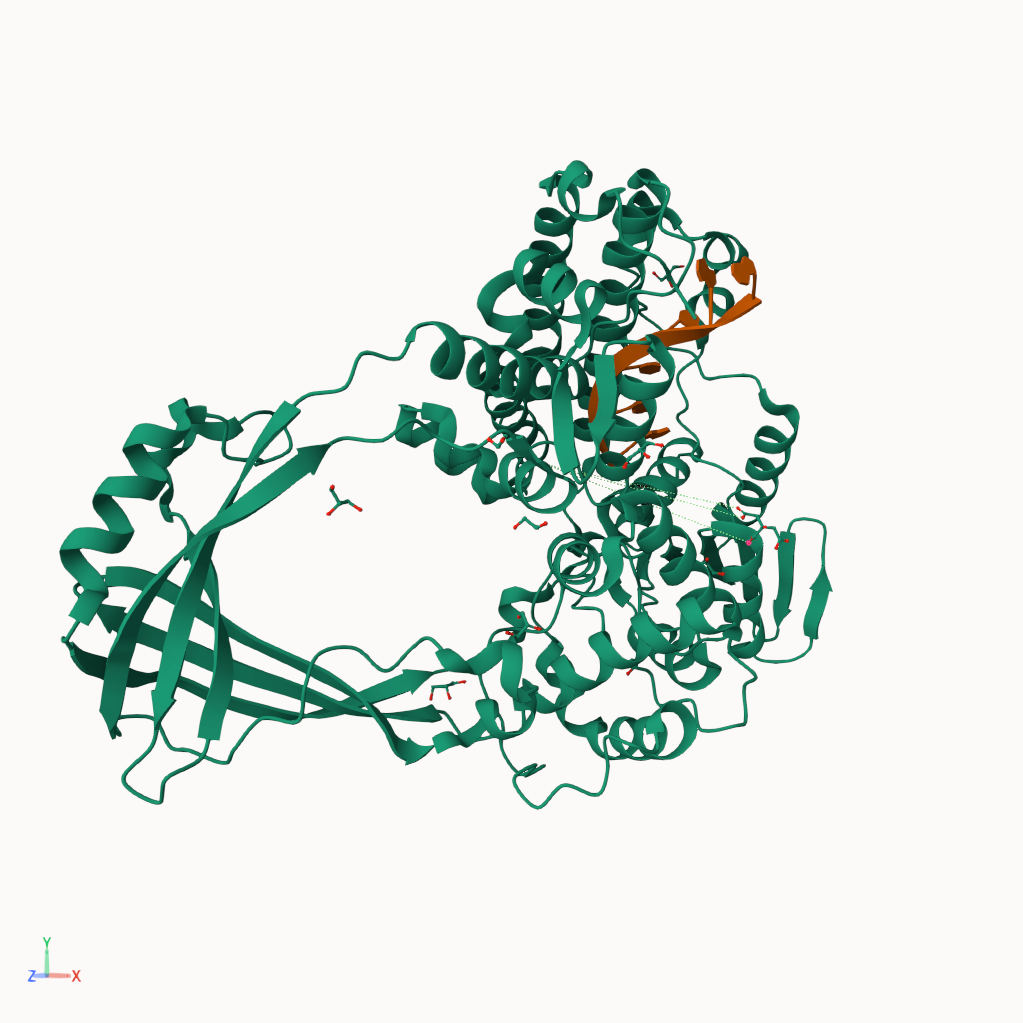}
}
\par\textbf{Biology}
\par Solvent hidden
\end{minipage}\hfill
\begin{minipage}[t]{0.19\linewidth}
\centering
\parbox[c][33mm][c]{\linewidth}{\centering
\includegraphics[width=\linewidth,height=33mm,keepaspectratio,trim=140bp 157bp 182bp 130bp,clip]{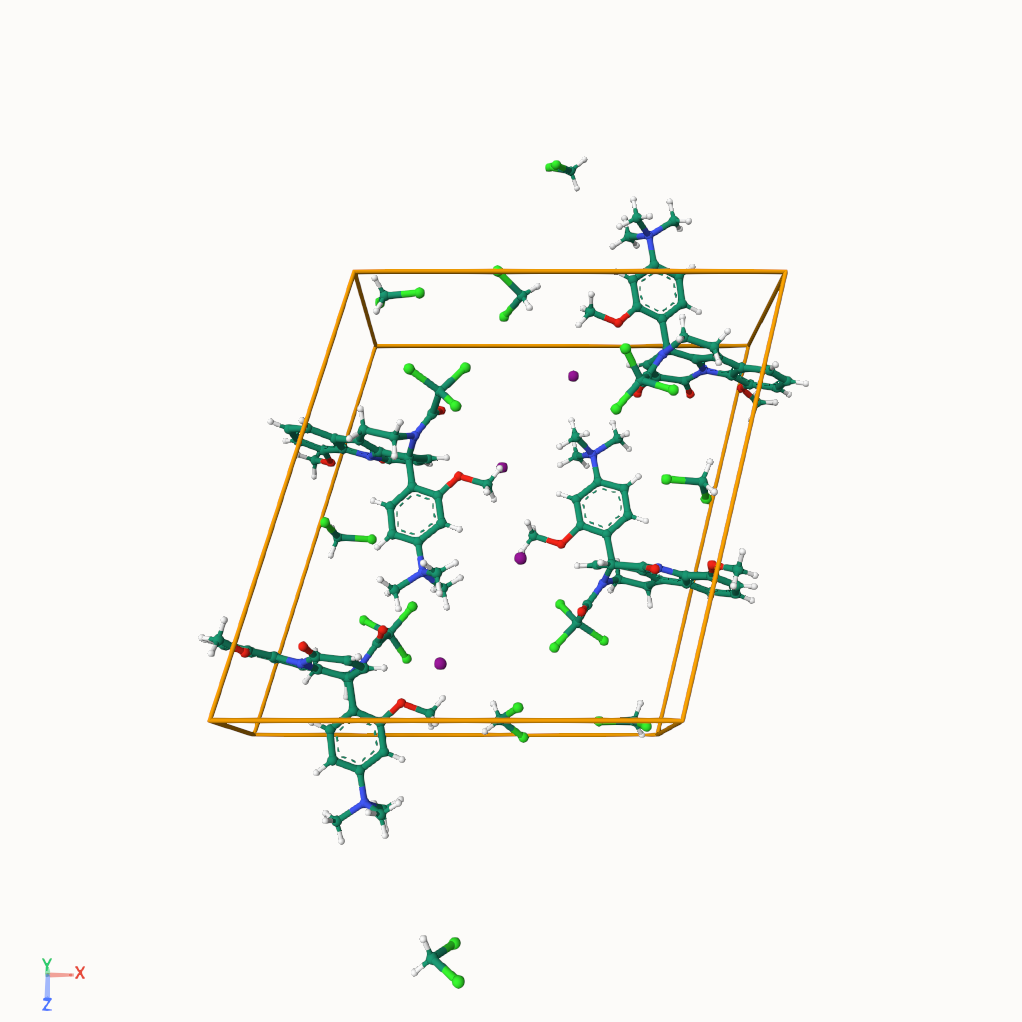}
}
\par\textbf{Chemistry}
\par Unit-cell view
\end{minipage}\hfill
\begin{minipage}[t]{0.19\linewidth}
\centering
\parbox[c][33mm][c]{\linewidth}{\centering
\includegraphics[width=\linewidth,height=33mm,keepaspectratio,trim=260bp 14bp 74bp 210bp,clip]{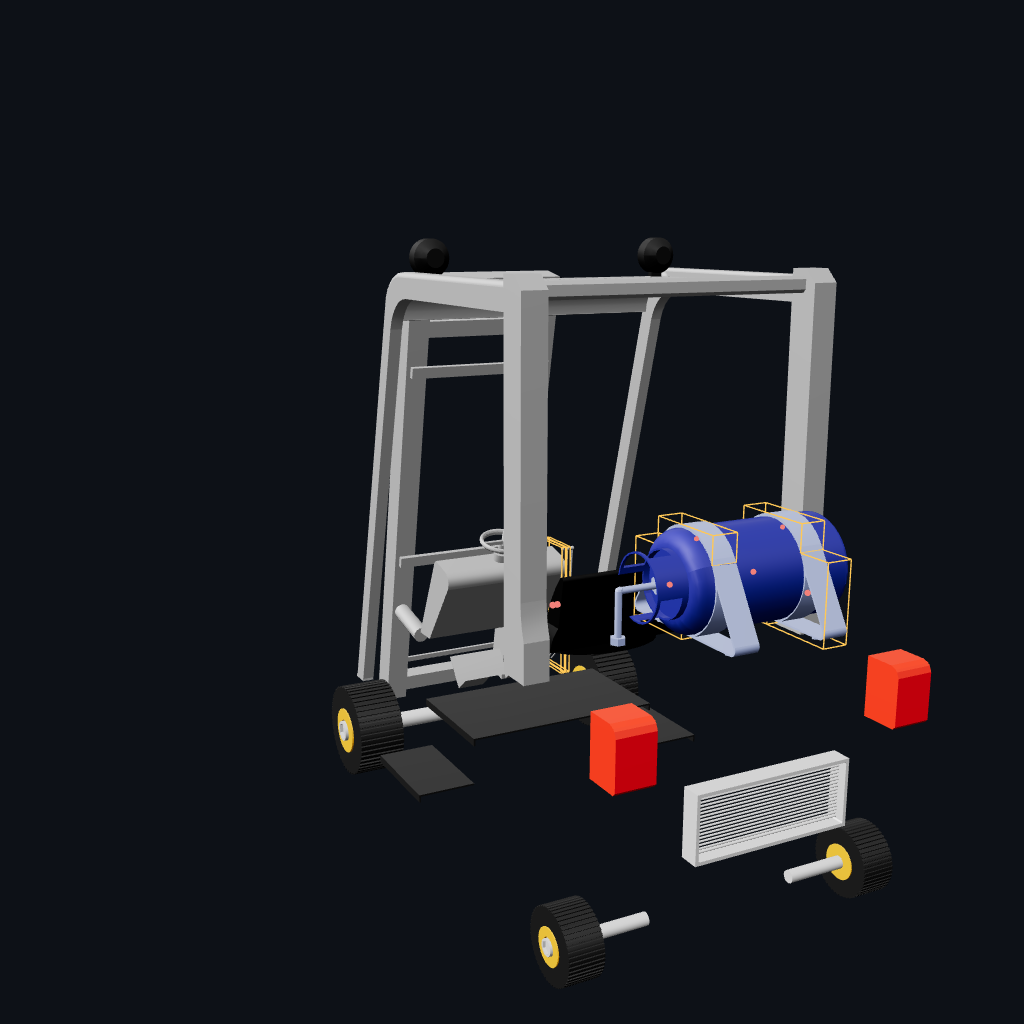}
}
\par\textbf{Mechanical CAD}
\par Body/fork hidden
\end{minipage}\hfill
\begin{minipage}[t]{0.19\linewidth}
\centering
\parbox[c][33mm][c]{\linewidth}{\centering
\includegraphics[width=\linewidth,height=33mm,keepaspectratio]{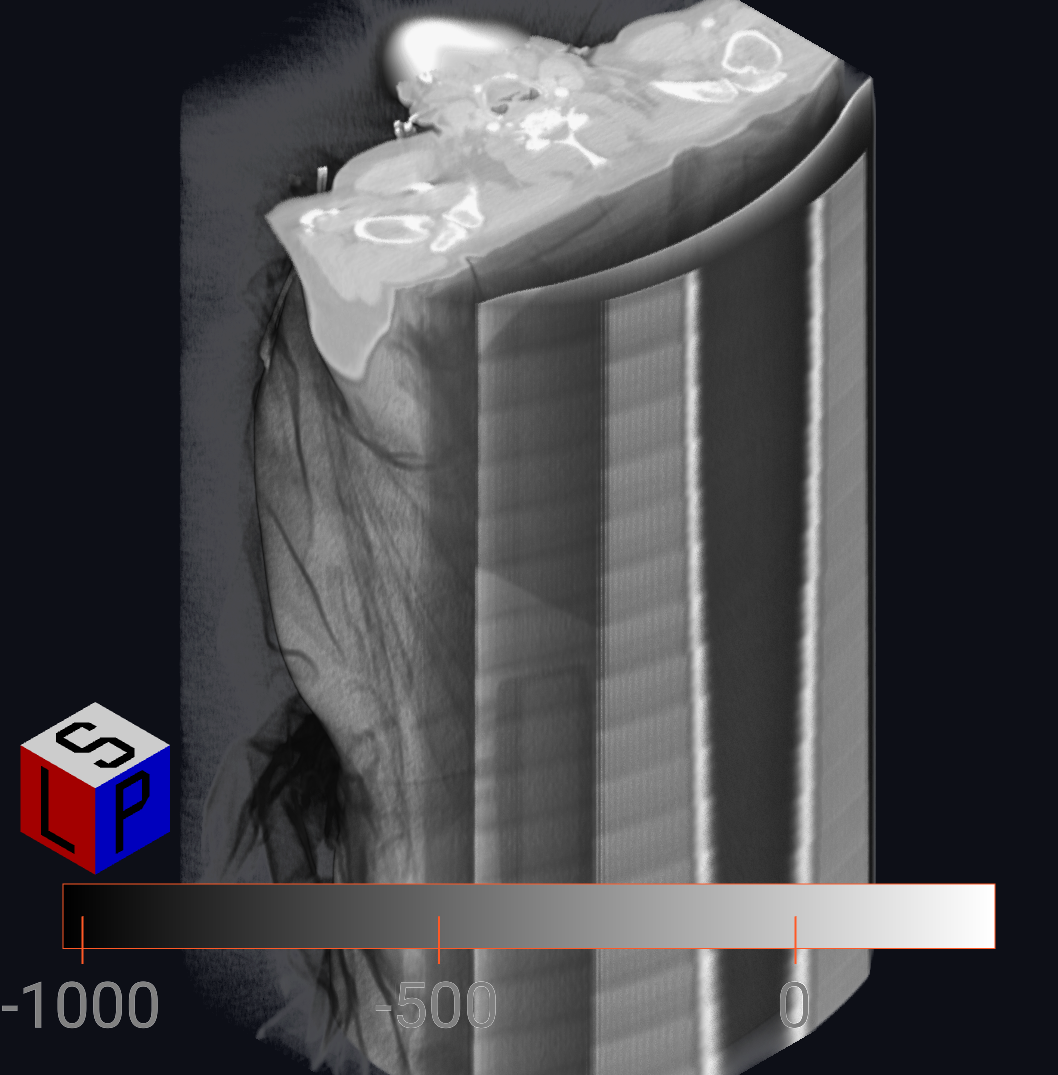}
}
\par\textbf{Medical Imaging}
\par 3D volume
\end{minipage}
\caption{\textbf{3D inspection across five scientific domains.} Left to right: architectural cutaway, protein after hiding water and ions, crystal unit cell, CAD assembly after hiding the body and fork, and CT volume rendering. The first four are recorded \href{https://unipat.ai/blog/ScienceVision}{public-demo states}; the medical panel is an additional 3D volume view of the same CT source.}
\label{fig:sciencevision-gallery}
\end{figure}

Figure~\ref{fig:sciencevision-gallery} shows inspected 3D structures across five domains, including a complementary CT volume view. Separate Chemistry and Medical Imaging examples in Figure~\ref{fig:dataset-cases} (Appendix~\ref{newapp:tasks}) illustrate how target visibility and measurement conventions enter the requested answers.

\subsection{Tasks and scientific evidence}
Tasks range from direct structural readings and counts to domain-specific judgments. Each pairs a native 3D object with a scientific question, answer requirements, and a reference answer. The question identifies a target through a scientific description or its appearance in a specified view. The answer requirements define the requested quantities or identities, units, coordinate frame, and conventions. Responses range from a single category to several related measurements or locations; answer entries need not coincide with independently scored components.

Scientific judgment enters through the relation between a target, its structure, and the requested conclusion. In the chemistry example in Figure~\ref{fig:dataset-cases}, initial-view occlusion identifies two atoms. Their identities select a shortest bonded path; the solver must return its intermediate atom IDs and the minimum absolute dihedral angle over consecutive four-atom segments. The stored path minimum is $64.49^{\circ}$. Choosing different atoms changes the scientific quantity being measured.

The medical example supplies the diagnostic target, a small left pleural effusion, and requests its extent along the three array axes, all longest axes, and a signed left--right centroid offset. The reference extent is $[86,81,61]$ voxels. Extents include both bounding voxels, whereas the centroid offset is a physical distance in millimetres. These examples illustrate how target identity and scientific conventions determine an answer; they do not define a common process followed by all tasks.

\subsection{Coverage}
The five equally represented domains contain 22 subdomains, with four to 25 tasks per subdomain. Architecture includes spatial continuity and dimensional judgments; Biology includes contacts and spatial relations; Chemistry includes conformation and molecular geometry; Mechanical CAD includes composition and component properties; Medical Imaging includes localization, slice comparison, and measurements. Equal domain quotas balance their aggregate weights, not the frequency of scientific questions in practice or intrinsic task difficulty.

Appendix~\ref{newapp:tasks} provides the complete subdomain distribution, answer organization, and five-domain examples. Reference inspections are available for all 250 tasks, with 2,156 recorded operations excluding the initial states. They document expert activity, rather than a minimum reasoning length or a human performance baseline.

\begin{figure}[!htbp]
\centering
\includegraphics[width=0.9\linewidth]{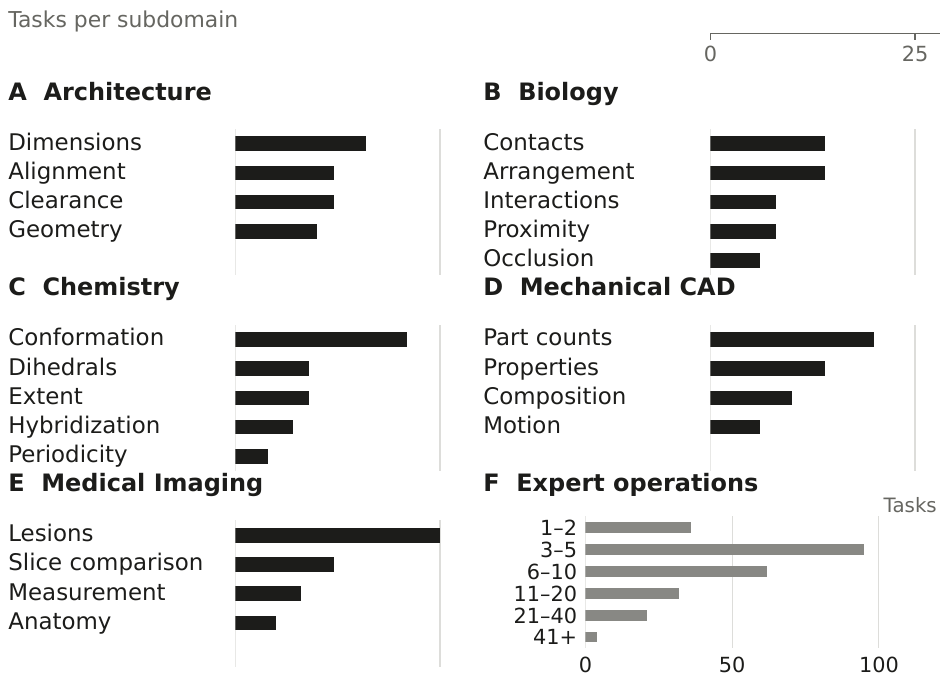}
\caption{Dataset coverage and expert reference inspections. Panels A--E use short labels for the 22 subdomains and a shared task-count scale; each domain contains 50 tasks. Panel F shows recorded expert operations per task, excluding the initial viewer state.}
\label{fig:dataset-coverage}
\label{fig:benchmark-profile}
\end{figure}

\subsection{Construction and reference validity}
\label{sec:benchmark-validity}
Objects are checked for question support and viewer access. Candidate questions are prepared from templates or curated drafts, sometimes with agent assistance. Experts review targets and domain assumptions, revise questions, inspect the objects, and supply reference answers. Quality review accepts the annotations or returns them for revision. Appendix~\ref{newsec:construction} preserves the detailed construction account.

The references comprise 110 recorded values, 106 script-derived values, and 34 tool-derived values. These categories describe provenance, not success rates for independently recomputing answers. The scoring configurations contain 54 task-specific adjustments, including component rubrics, aliases, and numerical tolerances. A reference value alone does not specify all accepted alternatives. Appendix~\ref{newapp:verification} separates construction, reference evidence, and scoring scope.

A prior independent audit found underspecified targets or counting units in eight CAD questions without establishing numerical reference errors. We retain those tasks in the reported benchmark and assess their influence separately in Section~\ref{sec:reliability}. File consistency, reference provenance, and a partial structural audit support different checks; none certifies the correctness of every reference.

\par
\ifdim\dimexpr\pagegoal-\pagetotal\relax<5\baselineskip
  \newpage
\fi
\section{Evaluation Design}
\label{newsec:protocol}

\subsection{Evaluation settings}
Models run in agent harnesses that manage file access, code execution, and tool interactions. In one setting, visual question answering uses native 3D artifacts accessed through the agent harness's file and code-execution tools (\textbf{Harness + 3D}), allowing geometric and structural calculations. The same workflow is augmented with MCP viewer inspection and measurement tools (\textbf{Harness + 3D + MCP}); tool use is optional, so the setting permits either computation or visual inspection. The third setting is visual question answering from prescribed rendered 2D images using an agent harness (\textbf{Harness + 2D}). We use these compact labels in tables, figures, and configuration comparisons. In the rendered-image setting, the main comparison uses six fixed camera positions and a fixed image order; agents cannot request another view. Medical tasks pair volume renders with slices, yielding twelve images for six nominal views. All settings use the same VQA questions, response requirements, and task-specific deterministic final-answer scoring rules.

The archived score exports label Harness + 3D and Harness + 3D + MCP as \texttt{Source} and \texttt{Source+MCP}, respectively.

These evaluation settings specify different evidence representations and operations, not identical visual inputs. Harness + 3D--Harness + 2D differences therefore do not isolate active observation. Complete access descriptions and the limits of run-specific tool verification appear in Appendix~\ref{newapp:access}.

\subsection{Models and execution configurations}
\label{sec:execution-main}
We evaluate the 12 models in Table~\ref{newtab:main} on the same 250 tasks under all three conditions. Each configuration combines a model, an agent harness, and evidence access. Codex is assigned to GPT-6 Astra and GPT-6.1 Sol, and Claude Code to the other ten models \citep{OpenAI2026Codex,Anthropic2026ClaudeCode}. GPT Codex uses maximum reasoning effort; Claude Code effort is not recorded. Study-level Harness + 2D harness assignments cannot be independently verified for every retained run.

The protocol specifies three repeats per task. For Gemini under Harness + 3D and Harness + 3D + MCP, the mean and min--max range use the first two complete 250-task runs ($n=2$), excluding the partial third run. Other configurations use the available scores from the three planned runs. Hunyuan has three tasks with two repeats under Harness + 2D. Earlier outputs were retained alongside declared reruns of 39 tasks after question or scoring changes. These records constrain comparisons across inputs, scorers, budgets, and model versions. Appendix~\ref{newapp:config} documents the available execution evidence; supplementary alternative-harness runs remain separate from the canonical leaderboard.

\subsection{Scoring and aggregation}
\label{sec:scoring-main}
Task-specific rules compare each submitted answer with its reference, assigning $s_{ir}\in[0,1]$ to task $i$ and repeat $r$. They specify scored components, matching, and partial credit, without necessarily averaging all stored answer entries. We apply no additional universal eligibility gate to exported scores (Appendix~\ref{newapp:verification}).

For the set $R_i$ of selected available scored repeats, the task mean is $\bar{s}_i=|R_i|^{-1}\sum_{r\in R_i}s_{ir}$. The mean over task set $D$ is $S_D=|D|^{-1}\sum_{i\in D}\bar{s}_i$, weighting tasks equally, with $|D|=50$ per domain and $250$ overall. Min--max endpoints are the lowest and highest complete-run means on the same 250 tasks. Missing repeats are omitted, recorded zeros retained, and intermediate values unrounded. These means retain partial credit and differ from complete-answer accuracy.

Cross-configuration differences describe the evaluated settings and do not establish a controlled MCP effect. Complete scores and available execution records appear in Appendices~\ref{newapp:full-results} and~\ref{newapp:config}, respectively.

\section{Empirical Findings}
\label{newsec:results}

\subsection{Performance across evaluation settings}
Figure~\ref{newfig:leaderboard} ranks the 12 canonical model--harness configurations under Harness + 3D. Figures~\ref{newfig:leaderboard-mcp} and~\ref{newfig:fixed6} in Appendix~\ref{newapp:canonical-results} rank Harness + 3D + MCP and Harness + 2D independently; Table~\ref{newtab:main} gives all 36 scores. GPT-6 Astra/Codex has the highest Harness + 3D score, $0.5291$, while Claude Fable~5.1 has the highest Harness + 2D score, $0.2965$. Harness + 3D exceeds Harness + 2D for every model by $7.73$--$28.49$ percentage points; Harness + 3D + MCP also exceeds Harness + 2D in every row. These are differences between evaluated configurations, including permitted source computation, rather than isolated benefits of visual inspection.

The strongest observed answer score in the main comparison remains about one half of the available credit. This score allows partial answers, so it does not imply that half the questions were answered completely. Appendix~\ref{newapp:canonical-results} retains the exact scores, configuration details, and the Harness + 2D leaderboard. The main plot and complete table use the same unrounded task means.

\subsection{Domain performance and MCP score differences}
Table~\ref{tab:main-domain-3d} breaks down the overall Harness + 3D leaderboard in Figure~\ref{newfig:leaderboard} by domain. Within each column, the 12 model--harness configurations are compared on the same 50 tasks.

\begin{table}[!htbp]
\centering
\caption{\textbf{Model comparison within each domain under Harness + 3D.} Each column compares models on the same 50 tasks from one domain. Harness + 3D gives models access to native 3D artifacts through file and code tools. Cells report scores from 0 to 1 (higher is better): selected available repeats are averaged per task, then tasks are weighted equally. Within each column, \textbf{bold} and \underline{underlined} values mark the highest and second-highest unrounded mean scores, respectively. GPT models use Codex; all other models use Claude Code.}
\label{tab:main-domain-3d}
\includegraphics[width=\linewidth]{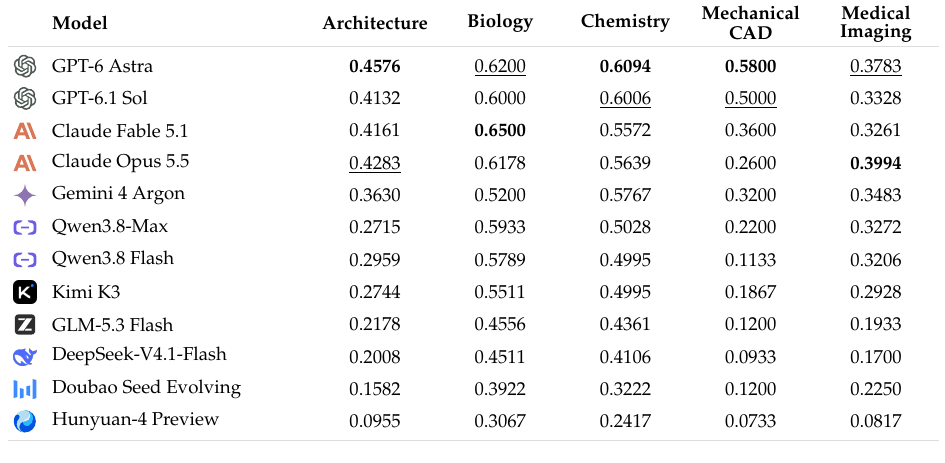}
\end{table}

GPT-6 Astra records the highest mean@3 in architecture, chemistry, and mechanical CAD. In biology, Claude Fable~5.1 scores 0.6500 compared with 0.6200 for Astra; in medical imaging, Claude Opus~5.5 scores 0.3994 compared with 0.3783 for Astra.

We next compare overall scores between Harness + 3D + MCP and Harness + 3D. These changes are smaller than the Harness + 3D--Harness + 2D differences and have no common direction. Eight configurations increase and four decrease, ranging from $-2.40$ percentage points for Kimi K3 to $+2.17$ points for GPT-6.1 Sol. Astra changes by $-0.08$ points.

Paired trajectories explain why these averages are small and unstable. In the inspected pairs, MCP helped when the interface exposed the quantity the scorer checks: one Harness + 3D + MCP trial measured the reference contact at $2.229$~\AA{} after the Harness + 3D trial excluded that pair as a direct link. It helped little when the interface did not cover the asset format, or when the returned value did not follow the scorer's sign and precision convention. Some paired differences were not tool effects at all: in the 94-task Fable control, a CAD pair computed the same extents and ratio but rounded $0.625$ to $0.62$ versus $0.63$, and fewer than one third of Harness + 3D + MCP trials called an MCP tool. Appendix~\ref{newsec:mcp-trajectories} gives the cases and call counts; they are descriptive comparisons, not estimates of a controlled MCP effect.

\begin{figure}[!htbp]
\centering
\includegraphics[width=\linewidth]{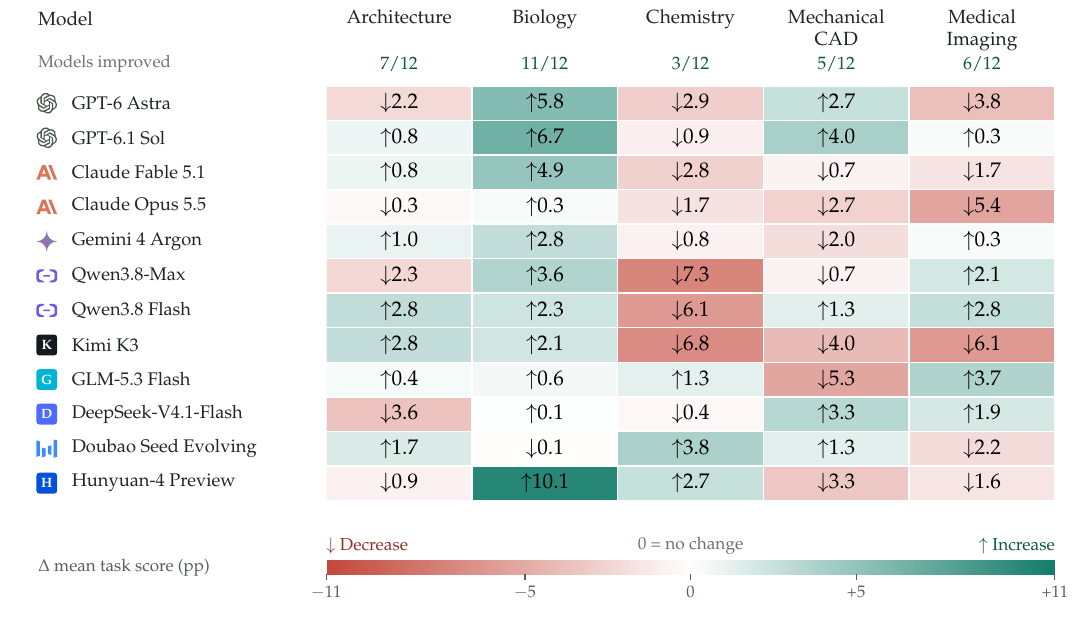}
\caption{\textbf{MCP score changes across domains.} Cells show Harness + 3D + MCP minus Harness + 3D in percentage points, with a shared color scale. Column headings report the number of models with higher scores.}
\label{newfig:domains}
\end{figure}

Comparing Harness + 3D with Harness + 3D + MCP, Biology scores rise for 11 of the 12 canonical configurations, with an equal-model mean difference of $+3.26$ percentage points; Chemistry scores fall for nine, with a mean difference of $-1.83$ points. Domain changes can also cancel within a model: Astra's Biology score rises from $0.6200$ to $0.6778$, while Medical Imaging falls from $0.3783$ to $0.3406$. Sol's Biology score rises from $0.6000$ to $0.6667$. Among the canonical configurations, Fable~5.1 with MCP reaches the highest domain mean, $0.6989$ in Biology. Figure~\ref{newfig:domains} makes these opposing domain changes visible. Figure~\ref{newfig:domains-fixed6} gives Harness + 2D domain scores; the domain subtotal rows in Appendix~\ref{app:subdomain-results} provide all 180 absolute domain means.

Task-family analysis sharpens this contrast. Equal-weight averages over the 12 canonical models show Harness + 3D--Harness + 2D gaps of 45.06 percentage points for dihedral measurement ($N=9$) and 40.68 points for conformation/occlusion ($N=21$). Every model scores higher with Harness + 3D in both families, connecting the aggregate result to the molecular-geometry judgments in Section~\ref{newsec:tasks}. These differences remain descriptive comparisons of available evidence and computation.

On 25 lesion-pointing tasks, corresponding means are 0.1011, 0.0967, and 0.0344 for Harness + 3D, Harness + 3D + MCP, and Harness + 2D. All 36 configurations score lower there than on slice comparison. Each model also has positive and negative MCP differences across the 22 subdomains. Thus, a single domain or overall mean does not describe all scientific judgments. Appendix~\ref{app:subdomain-results} gives the aggregation definitions and all 792 subdomain values; these small, selected families are not population-level ability estimates.

\subsection{Reliability and interpretation of the findings}
\label{sec:reliability}
Excluding the eight CAD questions with target or counting-unit ambiguity leaves 242 tasks. All three model rankings, every Harness + 3D--Harness + 2D direction, and the eight-increase/four-decrease MCP split remain unchanged. Astra Harness + 3D changes from $0.5291$ to $0.5356$, and Fable Harness + 2D from $0.2965$ to $0.3063$. This sensitivity check preserves the original benchmark and does not validate all remaining references; Appendix~\ref{newapp:full-results} gives the complete diagnostic.

Execution settings also affect the apparent score ceiling. Supplementary Opus/Codex Harness + 3D ($0.5448$) and Fable/Codex Harness + 3D + MCP ($0.5349$) exceed the canonical maxima. Codex scores are higher in the eight available full-set harness comparisons by $4.56$--$13.40$ percentage points, but execution settings and evaluation records differ. They do not establish a controlled harness effect or a universal ordering. Table~\ref{tab:ablation-evidence-harness} gives the full comparison; Appendix~\ref{newapp:harness-process} gives the recovery case.

More fixed views do not consistently improve scores in the earlier camera study: neither plan gains from six to twelve views (Table~\ref{tab:ablation-camera-core}). This historical pattern does not establish an optimal view budget for the current benchmark.

MCP-only inspection and Harness + 3D + MCP test distinct access protocols (Table~\ref{tab:ablation-evidence-mcp}). The historical tool-only record does not estimate the marginal benefit of adding MCP to source computation in the current evaluation.

Answer scores cannot identify whether a failure arose from target selection, evidence acquisition, domain inference, or output interpretation. Recorded expert inspections show an intended route rather than an evaluated model's observations. Complete task-score profiles and available trajectory evidence are retained in Appendix~\ref{newapp:full-results} and Appendix~\ref{newapp:harness-process}; they do not supply a human baseline or a validated process-error taxonomy.

\section{Related Work}
\label{newsec:related}

\paragraph{Scientific structural question answering.}
ScanQA grounds questions in indoor 3D scans, while SQA3D adds a described position and orientation for situated question answering \citep{azuma2022scanqa,ma2023sqa3d}.
DeepTumorVQA organizes questions about 3D CT into recognition, measurement, visual reasoning, and medical reasoning \citep{Chen2026DeepTumorVQA}.
ProtStructQA defines questions over AlphaFold-predicted proteins using formal task definitions for structural measurements and labels \citep{Mandiga2026ProtStructQA}.
Both benchmarks connect object-derived evidence to explicit answer criteria within a scientific domain.
\bench{} uses task-specific references and answer requirements across five scientific and engineering domains, under the relevant domain conventions.

\paragraph{Scientific visualization agents and evidence access.}
SciVisAgentBench evaluates scientific analysis and visualization workflows, including analytical question answering and spatial outputs \citep{Ai2026SciVisAgentBench}.
It grades outcomes against expert-authored rubrics using multimodal judges and deterministic evaluators.
\bench{} grades answers to scientific questions about supplied artifacts using task-specific deterministic checks without an LLM judge.
The distinction concerns task objectives and graded outcomes; question answering and interaction are already present in prior scientific 3D evaluations.

Scientific agents obtain evidence through generated analysis scripts \citep{chen2024scienceagentbench,Peterka2025ChatVis}, visual feedback, and direct viewer operations \citep{Liu2024AVA,Liu2025ParaViewMCP}.
MedOpenClaw lets agents operate medical viewers, including 3D Slicer; MedFlowBench checks both task answers and submitted spatial evidence \citep{Shen2026MedOpenClaw}.
\bench{} compares native-artifact access with or without MCP inspection tools and rendered 2D images under the same final-answer scoring rules.
These settings distinguish available evidence and operations from the answer being graded.
Tool availability alone does not establish how an answer was obtained.
Table~\ref{tab:related-work-comparison} compares benchmark scope, question-answering objectives, evidence access, and grading. Appendix~\ref{newapp:related-comparison} gives additional scoring details.

\begin{table}[!htbp]
\captionsetup{font=small}
\caption{Comparison of benchmark scope, question-answering objectives, evidence access, and grading.}
\label{tab:related-work-comparison}
\centering
\footnotesize
\setlength{\tabcolsep}{3.5pt}
\renewcommand{\arraystretch}{1.08}
\begin{tabularx}{\linewidth}{@{}>{\raggedright\arraybackslash}p{0.28\linewidth}*{5}{>{\centering\arraybackslash}X}@{}}
\toprule
\textbf{Benchmark}
& \textbf{Cross-domain}\newline\textbf{science}
& \textbf{QA}\newline\textbf{focus}\textsuperscript{a}
& \textbf{Visual}\newline\textbf{interaction}\textsuperscript{b}
& \textbf{Static-image}\newline\textbf{QA}
& \textbf{Deterministic}\newline\textbf{scoring}\textsuperscript{c} \\
\midrule
DeepTumorVQA\newline\citep{Chen2026DeepTumorVQA}
& $\times$ & $\checkmark$ & $\checkmark$ & $\checkmark$ & $\checkmark$ \\
ProtStructQA\newline\citep{Mandiga2026ProtStructQA}
& $\times$ & $\checkmark$ & $\times$ & $\times$ & $\checkmark$ \\
OmniCAD\newline\citep{Wang2026OmniCAD}
& $\times$ & $\times$ & $\checkmark$ & $\times$ & $\checkmark$ \\
SciVisAgentBench\newline\citep{Ai2026SciVisAgentBench}
& $\checkmark$ & Partial & $\checkmark$ & $\times$ & Partial \\
GLUE3D\newline\citep{mariani2026glue3d}
& $\times$ & $\checkmark$ & $\times$ & $\checkmark$ & Partial \\
\midrule
\textbf{\bench{} (ours)}
& $\boldsymbol{\checkmark}$ & $\boldsymbol{\checkmark}$ & $\boldsymbol{\checkmark}$ & $\boldsymbol{\checkmark}$ & $\boldsymbol{\checkmark}$ \\
\bottomrule
\end{tabularx}
\par\vspace{3pt}
\begin{minipage}{\linewidth}
\footnotesize\raggedright
\textit{Notes:} QA, question answering; LLM, large language model; $\checkmark$, reported; $\times$, not reported. Cross-domain science spans multiple scientific fields. Static-image QA denotes a separate fixed-image evaluation.
\par\vspace{1pt}
\textsuperscript{a}\,QA focus follows the authors' task classification, including captioning when classified as QA. Partial denotes QA alongside visualization or processing tasks.
\par\vspace{1pt}
\textsuperscript{b}\,Visual interaction includes agent-selected views, slices, crops, or rendered prediction feedback.
\par\vspace{1pt}
\textsuperscript{c}\,Deterministic scoring concerns formal task outcomes: $\checkmark$, rule-based scoring without LLM judges; Partial, mixed rule-based and LLM grading.
\end{minipage}
\end{table}

\section{Discussion and Conclusion}
\label{newsec:discussion}
\label{newsec:conclusion}

ScienceVision evaluates scientific judgments about native 3D objects, not research-question formulation, experimental design, or validation of new scientific claims. Harness + 3D outperforms Harness + 2D for every tested model; MCP comparisons show smaller, bidirectional changes.

Equal domain quotas do not reflect scientific question frequencies; subdomains and object families remain uneven. Because selection used prior model performance, scores do not estimate difficulty for randomly sampled scientific questions. Reference inspections and annotation effort do not establish a human performance baseline. Reference answers may be incorrect; accepted-answer rules affect scores. Execution differences, incomplete repeats, and uncertain earlier runs limit capability attribution.

Even with native 3D objects and code execution, current models still struggle to make reliable scientific judgments.

\ificlrfinal
\section*{Contributors}

\textbf{Core contributors:} Tianyi Ma\textsuperscript{1,*}, Jing Gu\textsuperscript{2,*}, Haozhe Zhao\textsuperscript{1,\ddag}, Liang Chen\textsuperscript{1,\ddag}, Yunfan Zhang\textsuperscript{1}, Haijun Su\textsuperscript{2}, Jialong Wu\textsuperscript{1}, Le Xue\textsuperscript{2}, Arnav Das\textsuperscript{2}, Dustin Tran\textsuperscript{2}, Yinfei Yang\textsuperscript{2}, Andrew Dai\textsuperscript{2,\dag}, Kuan Li\textsuperscript{1,\dag}.

\textbf{Contributors:} Liangkun He\textsuperscript{1}, Zihao Guo\textsuperscript{1}, Xiang Gong\textsuperscript{1}, Yutao Zhou\textsuperscript{1}, Jiarui Sun\textsuperscript{1}, Chengda Wu\textsuperscript{1}, Zhao Zhang\textsuperscript{1}, Qianhui Yang\textsuperscript{1}, Zhenji Kang\textsuperscript{1}, Yi Shen\textsuperscript{1}, Wenda Wang\textsuperscript{2}.

\textbf{Affiliations:} \textsuperscript{1}UniPat AI.\quad\textsuperscript{2}Elorian AI.

\begingroup
\renewcommand{\thefootnote}{}
\footnotetext[0]{\textsuperscript{*}Equal contribution.\quad\textsuperscript{\ddag}Project leader.\quad\textsuperscript{\dag}Corresponding author.}
\endgroup
\fi

\label{sec:main-text-end}

\clearpage
\bibliography{bibliography/references}
\bibliographystyle{template/sciencevision}

\clearpage
\appendix
\section{Task Specifications, Coverage, and Context}
\label{newapp:tasks}

\subsection{Subdomain inventory}
Table~\ref{tab:dataset-subdomains} lists the 22 subdomains and their task counts. Each task belongs to one subdomain, although a question may require several kinds of observation or judgment. For example, a conformation question may also ask about an occluded atom. These groups describe task content rather than mutually exclusive scientific abilities. Supporting metadata records the assignment of each task.

\begin{table}[!htbp]
\caption{Complete subdomain composition. Objectives are illustrative descriptions of question content; they are not additional per-task ability annotations.}
\label{tab:dataset-subdomains}
\centering
\small
\setlength{\tabcolsep}{4pt}
\renewcommand{\arraystretch}{1.12}
\begin{tabularx}{\linewidth}{@{}l r >{\raggedright\arraybackslash}X@{}}
\toprule
Subdomain & Tasks & Illustrative objective \\
\midrule
\multicolumn{3}{@{}l}{\textbf{Architecture (50)}} \\
Dimensional compliance & 16 & Check dimensions against a stated requirement. \\
Alignment and stacking & 12 & Identify continuity or alignment of members. \\
Interference and clearance & 12 & Distinguish spatial interference from clearance. \\
Object geometry readings & 10 & Read object dimensions, arrangement, or scene geometry. \\
\addlinespace[3pt]
\multicolumn{3}{@{}l}{\textbf{Biology (50)}} \\
Molecular contact & 14 & Identify atom contacts and associated distances. \\
Spatial arrangement & 14 & Describe the relative arrangement of components. \\
Interaction typing & 8 & Identify interaction classes or coordinating atoms. \\
Proximity choice & 8 & Select the closest specified component or atom. \\
Occluded target geometry & 6 & Locate a view-defined target and assess its geometry. \\
\addlinespace[3pt]
\multicolumn{3}{@{}l}{\textbf{Chemistry (50)}} \\
Conformation and occlusion & 21 & Combine molecular conformation with occluded targets. \\
Dihedral measurement & 9 & Measure a specified torsion or its minimum value. \\
Molecular extent & 9 & Determine the separation of farthest heavy atoms. \\
Hybridization state & 7 & Classify and count heavy-atom hybridization states. \\
Crystal periodicity & 4 & Analyze a unit cell and its periodic contents. \\
\addlinespace[3pt]
\multicolumn{3}{@{}l}{\textbf{Mechanical CAD (50)}} \\
Part-instance counting & 20 & Count component copies, including hidden instances. \\
Part-property identification & 14 & Determine a part's geometric properties. \\
Assembly-composition counting & 10 & Count repeated parts or designs within an assembly. \\
Mechanism motion & 6 & Analyze gears, mechanisms, or assembly relations. \\
\addlinespace[3pt]
\multicolumn{3}{@{}l}{\textbf{Medical Imaging (50)}} \\
Lesion pointing & 25 & Locate a specified abnormality in a CT volume. \\
Slice comparison & 12 & Compare visible anatomy across specified slices. \\
Geometry measurement & 8 & Measure a structure's extent or position. \\
Anatomy pointing & 5 & Locate a named anatomical structure in a volume. \\
\midrule
\textbf{Total} & \textbf{250} & \textbf{22 subdomains} \\
\bottomrule
\end{tabularx}
\end{table}

\subsection{Answer structure and conventions}
The reference answers contain 143 single-entry responses and 107 responses with multiple recorded entries. There are respectively 143, 35, 45, 11, 5, 8, 2, and 1 references with 1, 2, 3, 4, 5, 6, 7, and 9 entries. A list counts as one entry, as does a single-word response. These counts describe answer organization and need not match the number of independently requested judgments.

Tasks can request an atom identity, an ordered list, a category, a coordinate, or a measurement. The question defines the convention. Voxel extents need not use physical distance units, and a list of intermediate atoms may exclude its endpoints. The examples below preserve the recorded reference values, without defining all tolerances or equivalent accepted answers.

\subsection{Reference inspection operations}
All 250 tasks have a recorded expert inspection, comprising 2,156 tool operations in total. We count viewer operations and exclude the initial state. The ranges 1--2, 3--5, 6--10, 11--20, 21--40, and 41 or more operations contain 36, 95, 62, 32, 21, and 4 tasks, respectively. These counts describe recorded expert activity, rather than model tool use or a minimum reasoning length.

\subsection{Chemistry example}
\label{newapp:chemistry-example}
The molecular-conformation question first asks for the smallest absolute dihedral angle formed by four consecutively bonded atoms with element sequence H-C-O-C. It then identifies two atoms by their occlusion in the initial view. The solver must report those atoms, the ordered intermediate atoms on their shortest bonded path, and the number of intermediate atoms. If the path has at least four atoms, the solver also reports its smallest absolute dihedral angle over consecutive four-atom segments. Otherwise, no path angle is reported.

\begin{figure}[!htbp]
\centering
\begin{subfigure}[t]{0.48\linewidth}
\centering
\includegraphics[width=52mm,height=52mm,keepaspectratio]{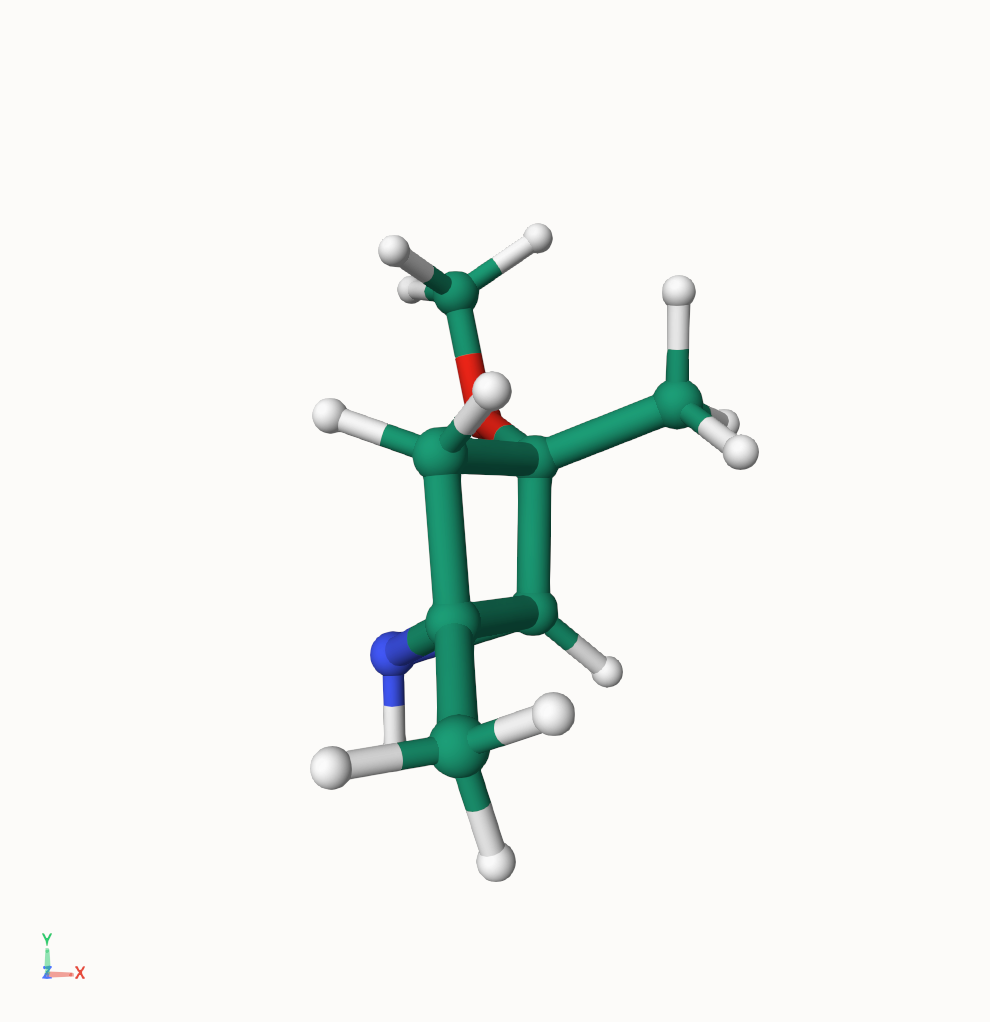}
\caption{Chemistry: initial view.}
\end{subfigure}\hfill
\begin{subfigure}[t]{0.48\linewidth}
\centering
\includegraphics[width=52mm,height=52mm,keepaspectratio]{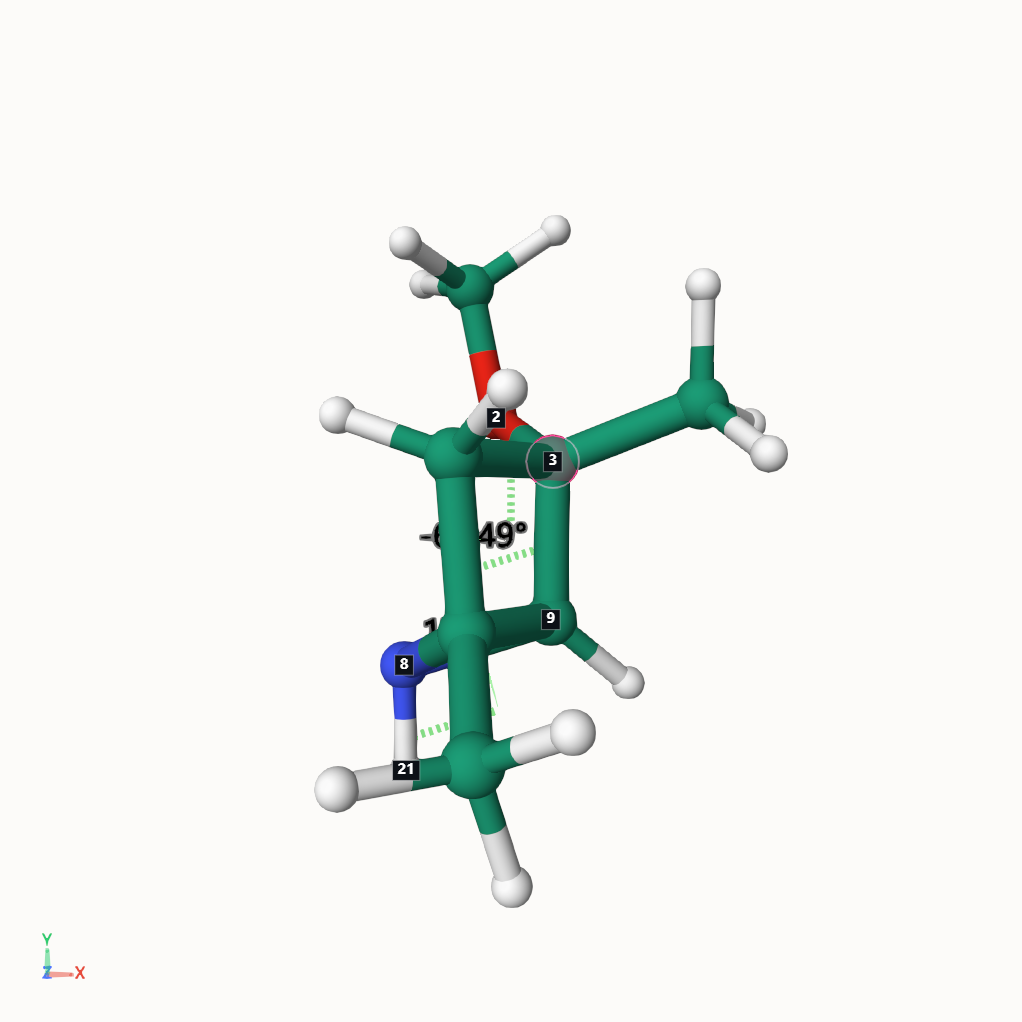}
\caption{Chemistry: labels and measurements.}
\end{subfigure}
\par\vspace{2mm}
\begin{subfigure}[t]{0.48\linewidth}
\centering
\includegraphics[width=65mm,height=65mm,keepaspectratio]{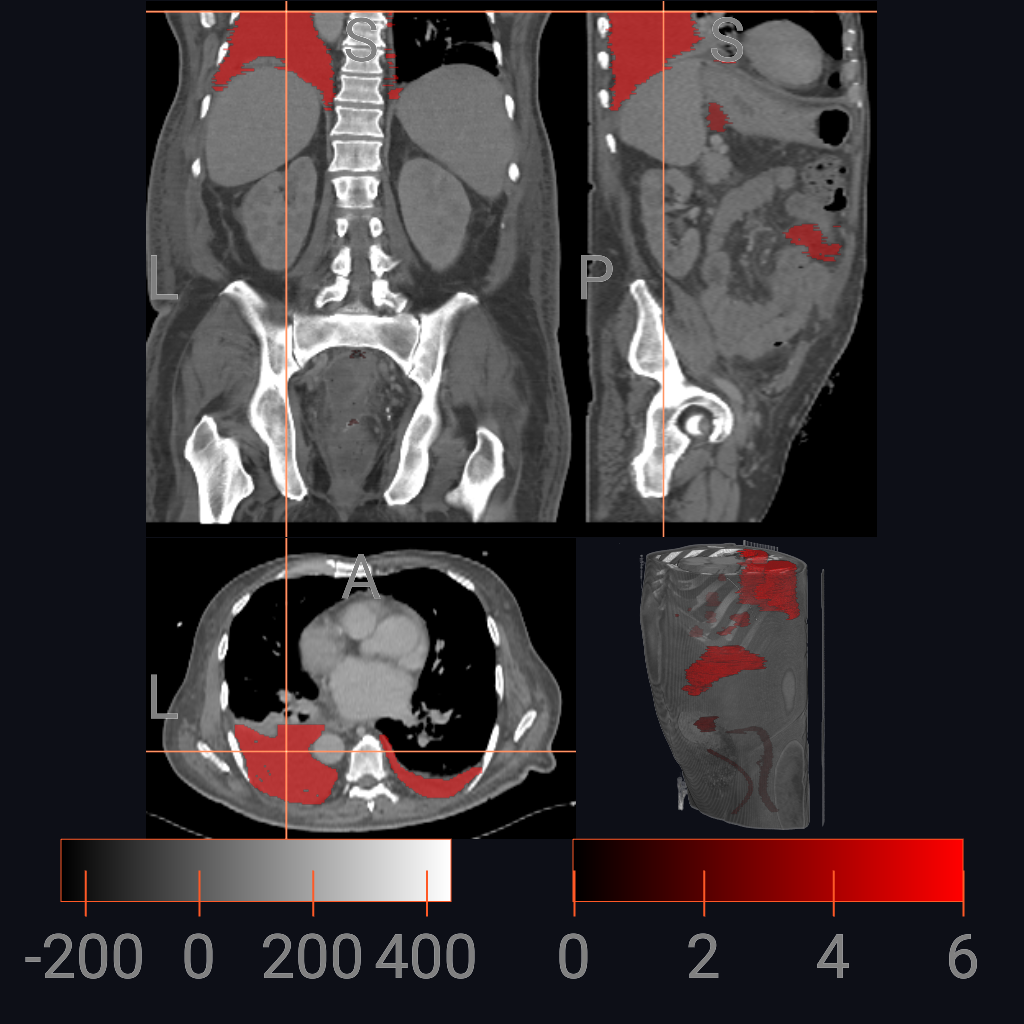}
\caption{Medical Imaging: original CT inspection.}
\end{subfigure}\hfill
\begin{subfigure}[t]{0.48\linewidth}
\centering
\includegraphics[width=65mm,height=65mm,keepaspectratio,trim=600bp 188bp 136bp 540bp,clip]{figures/assets/dataset-examples/medical.png}
\caption{Medical Imaging: 3D viewport enlarged.}
\end{subfigure}
\caption{Original recorded inspection states for the Chemistry (a,b) and Medical Imaging (c) examples. Panel (d) enlarges the 3D volume viewport within the same recorded medical frame shown in (c); it is not a separate operation.}
\label{fig:dataset-cases}
\end{figure}

The reference gives $53.34^{\circ}$ for the minimum H-C-O-C angle. The nearly fully occluded atom is 21, and the partially occluded heavy atom is 2. Their path contains three intermediate atoms, ordered 8, 9, and 3, with a minimum absolute path dihedral angle of $64.49^{\circ}$. These two angles answer different parts of the question. Figure~\ref{fig:dataset-cases} shows the initial view in panel (a) and the later labeled inspection in panel (b).

\subsection{Medical Imaging example}
The CT question identifies a small left pleural effusion and asks the solver to locate its centroid and representative slice. It also requests the target's extent along the three sampling-grid axes, every axis with the greatest extent, and the signed left--right centroid offset. Extents include both bounding voxels, and tied longest axes must be listed in increasing order. Measurements use at most six decimal places.

The recorded reference gives extents of $[86,81,61]$ voxels, identifies axis 0 as the longest, and gives a signed offset of $-64.56$ millimetres. Axis indices identify the sampling grid, rather than anatomical directions. Voxel extents should not be interpreted as physical lengths without the corresponding coordinate definition.

\subsection{Examples from architecture, biology, and mechanical CAD}
\begin{figure}[!htbp]
\centering
\begin{subfigure}[t]{0.48\linewidth}
\centering
\includegraphics[width=52mm,height=52mm,keepaspectratio]{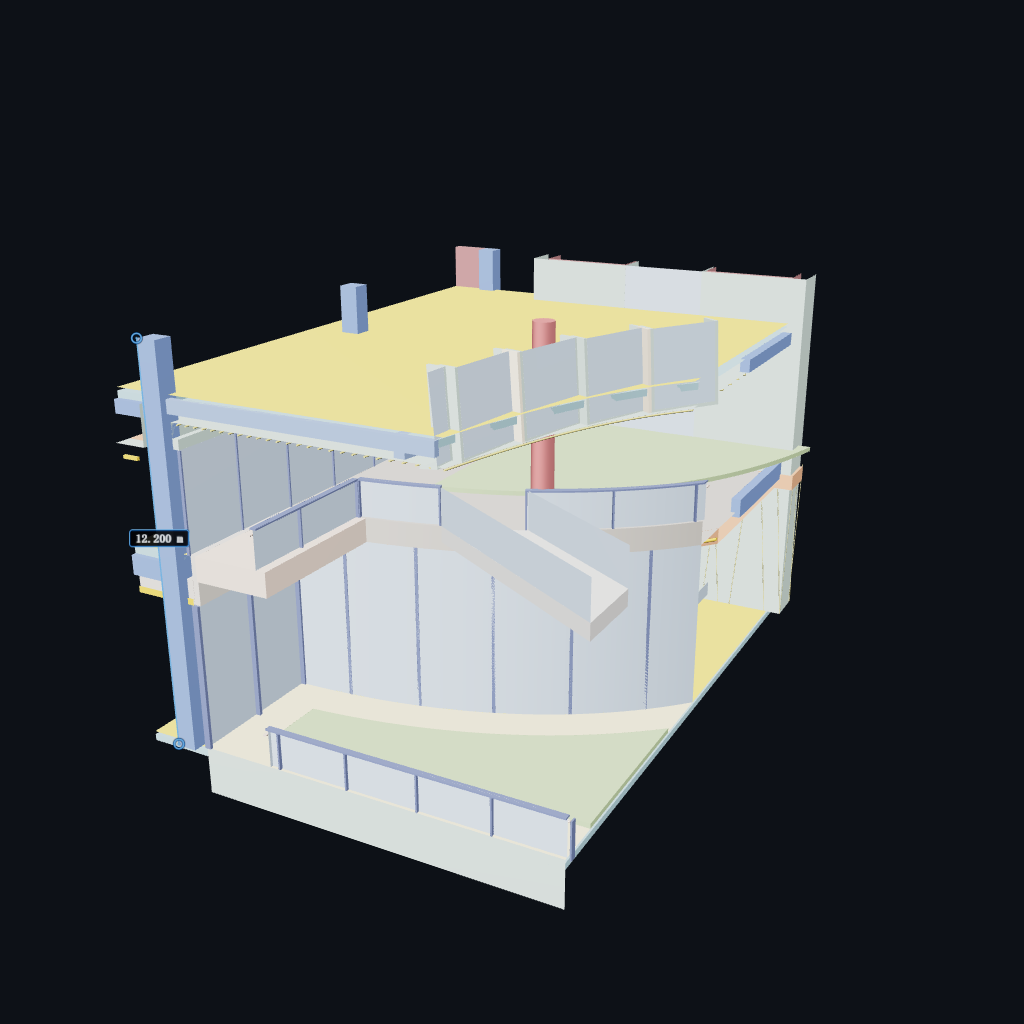}
\caption{Architecture.}
\end{subfigure}\hfill
\begin{subfigure}[t]{0.48\linewidth}
\centering
\includegraphics[width=52mm,height=52mm,keepaspectratio]{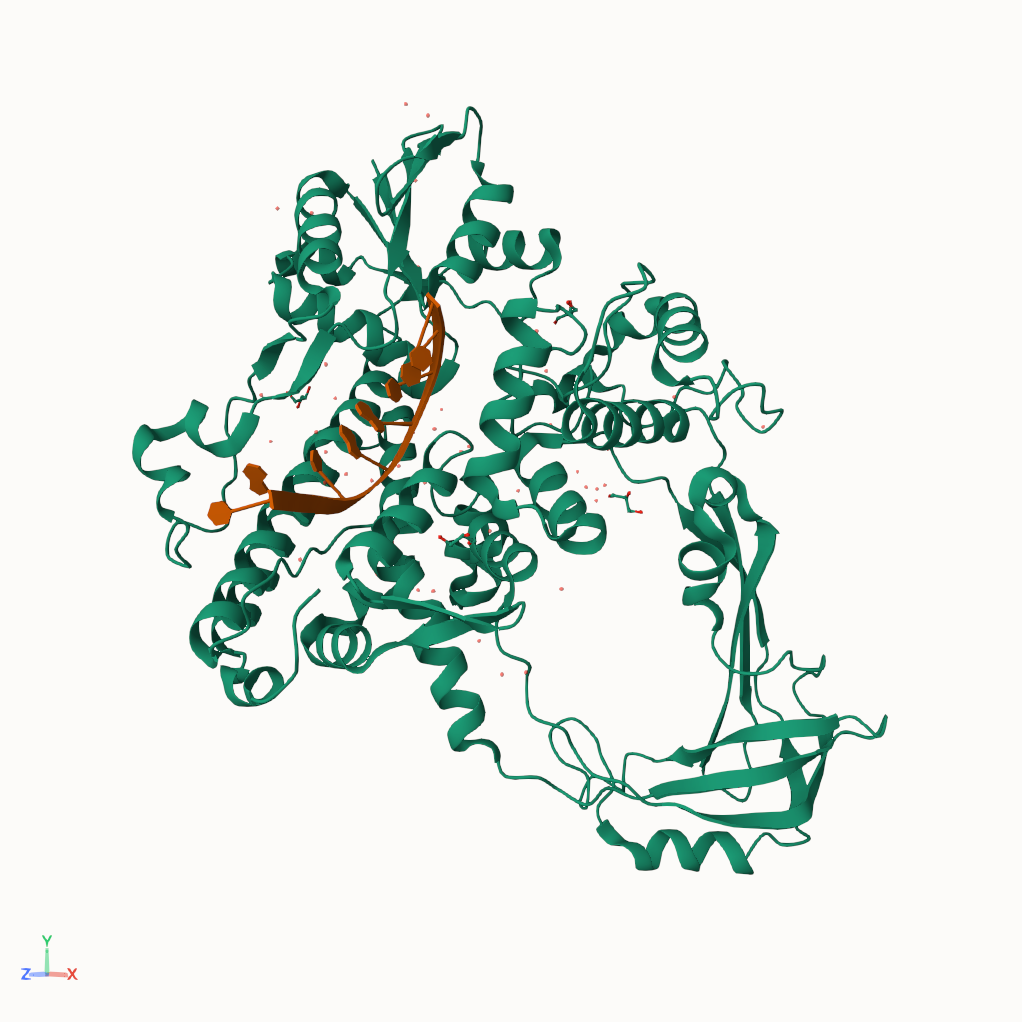}
\caption{Biology.}
\end{subfigure}
\par\vspace{2mm}
\begin{subfigure}[t]{0.48\linewidth}
\centering
\includegraphics[width=65mm,height=65mm,keepaspectratio]{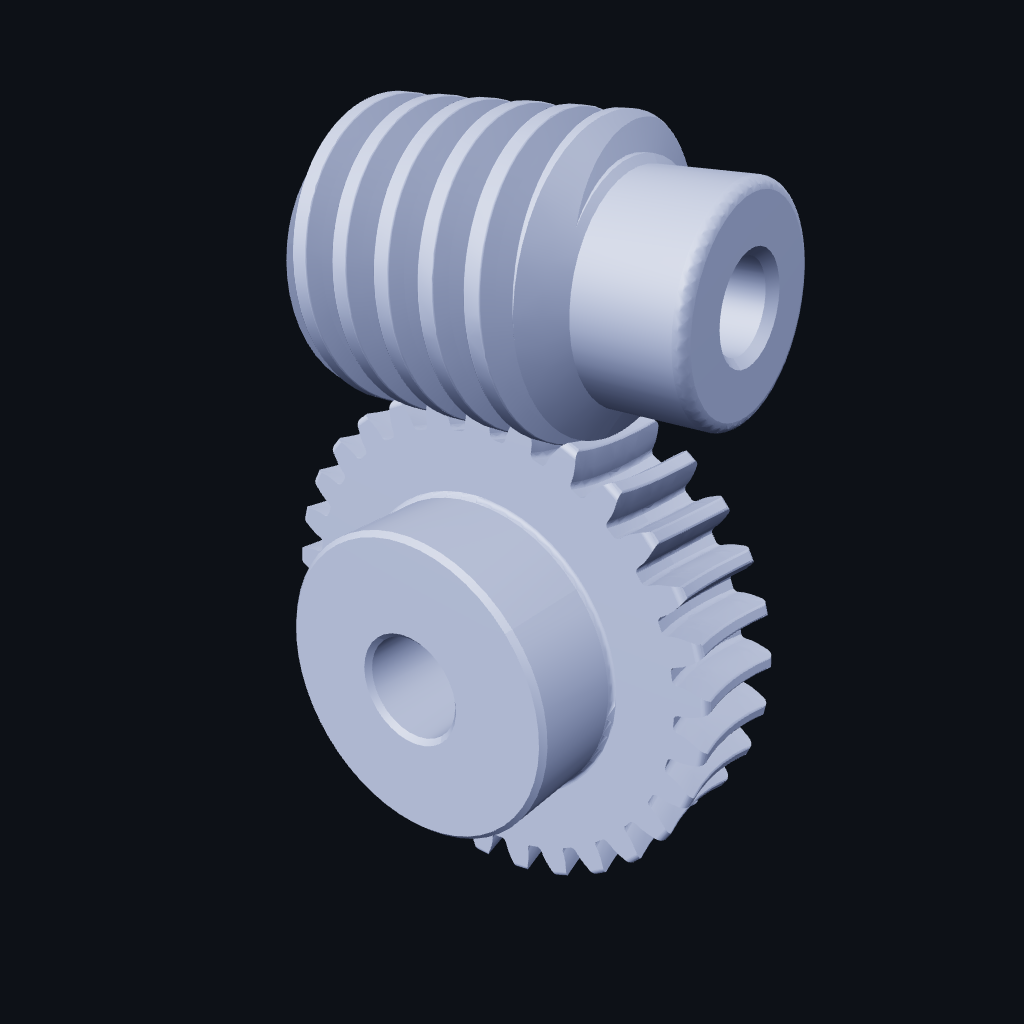}
\caption{CAD: initial 3D assembly.}
\end{subfigure}\hfill
\begin{subfigure}[t]{0.48\linewidth}
\centering
\includegraphics[width=65mm,height=65mm,keepaspectratio]{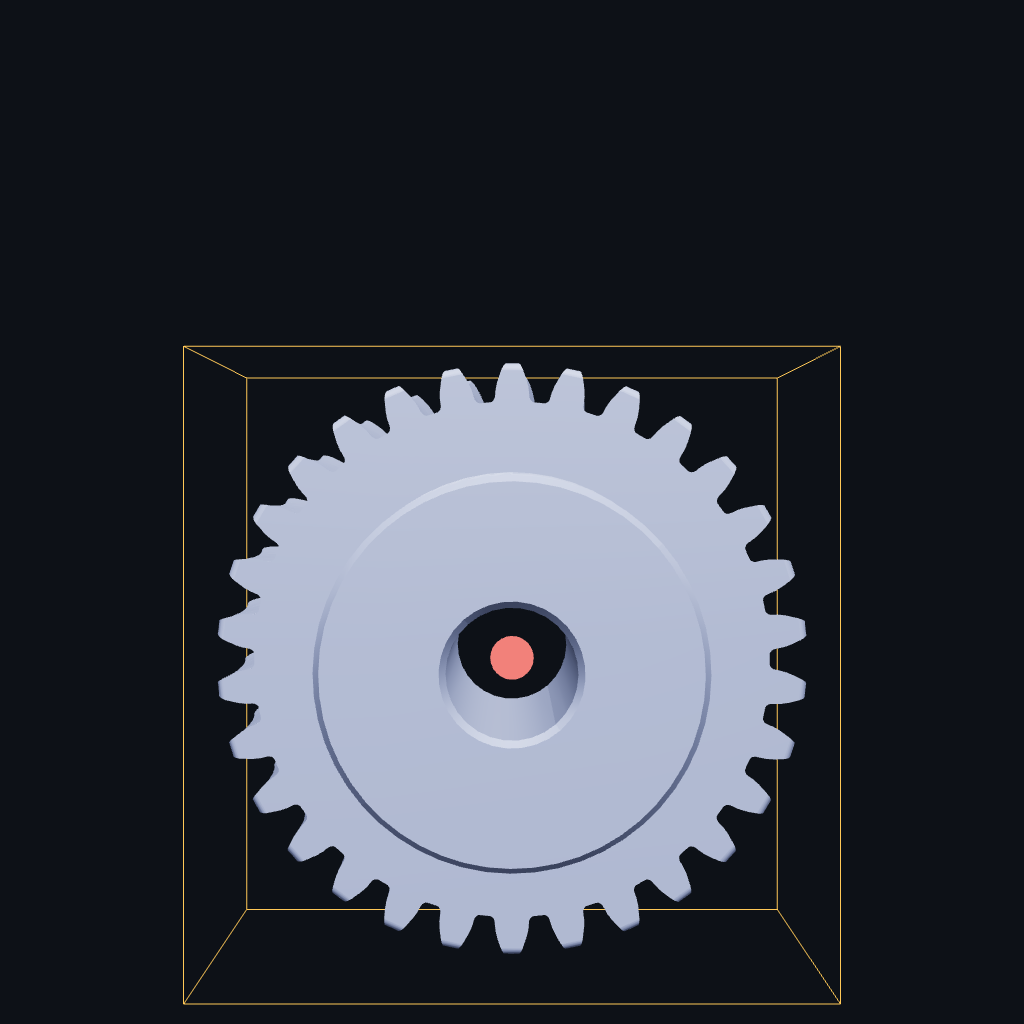}
\caption{CAD: worm hidden; front view.}
\end{subfigure}
\caption{Recorded reference inspections for architecture, biology, and mechanical CAD. Panels (c,d) show two states of the same worm--wheel assembly: the initial 3D view, followed by selecting the wheel, hiding the worm, and switching to the front view.}
\label{fig:dataset-other-cases}
\end{figure}

\paragraph{Architecture: continuity across storeys.}
The question asks which vertical structural members extend continuously across at least two storeys. It requests their count, shortest and longest vertical extents, and whether discontinuous members exist. Coordinates use millimetres with the vertical direction as the third axis. The viewer instead uses metres with the vertical direction as its second axis. Converting between them swaps the second and third coordinates, negates the new second coordinate, and multiplies all coordinates by 1,000.

The reference identifies four continuous members, with shortest and longest extents of 12,000 and 12,200 millimetres, respectively. It reports no discontinuous members.

\paragraph{Biology: terminal residues and chain position.}
The question asks whether the N-terminus or C-terminus of the largest polymer chain is closer to its centre. Chain size is measured by atom count, with ties resolved by the earliest chain identifier. The centre is the alpha carbon of the middle residue, choosing the later middle residue when the chain length is even. Terminal positions use the alpha carbons of the first and last residues in file order. The task uses the first structural model and the highest-occupancy alternate positions, excluding hydrogens and solvent. The reference answer is C-terminus.

\paragraph{Mechanical CAD: inspecting a gear assembly.}
The gear assembly is first shown in an oblique 3D view. The recorded inspection selects the wheel, hides the worm, and switches to a front view, exposing the wheel's perimeter (Figure~\ref{fig:dataset-other-cases}c,d). The task asks for the worm wheel's tooth count; the recorded reference is 32.

\subsection{Coverage and recorded inspection distribution}
Figure~\ref{fig:dataset-coverage} in the main text visualizes the subdomain composition and recorded inspection lengths; the exact counts and definitions are given above.

\subsection{Extended task-design description}
\providecommand{\sciencevisionDatasetAppendix}{newapp:tasks}

ScienceVision asks for structural and spatial judgments about specific 3D scientific objects. Its 250 tasks are evenly distributed across Architecture, Biology, Chemistry, Mechanical CAD, and Medical Imaging (Figure~\ref{fig:dataset-coverage}). The questions concern properties of particular objects, such as continuity across storeys in a building, contact between macromolecular structures, or the shape and multiplicity of assembly components.

\subsubsection{Objects, targets, and answer requirements}
Each task pairs an object with a question, answer requirements, and a reference answer. The question can identify its target by a scientific description or by its appearance in a specified view. The requirements specify what to report and define the relevant units, coordinates, and object-identity conventions. Responses range from a single category to several related quantities or locations. Expert inspection records preserve the viewer operations and visual states associated with the reference annotations; Appendix~\ref{\sciencevisionDatasetAppendix} details both the records and answer formats.

\subsubsection{From target selection to scientific judgment}
\paragraph{Chemistry: a view-defined target and its bonded geometry.}
In the chemistry example (Figure~\ref{fig:dataset-cases}a,b), the initial camera view defines two atoms by their occlusion. Their identities select the endpoints of a shortest bonded path. The task requests the intermediate atom IDs, excluding both endpoints, and the smallest absolute dihedral angle over consecutive four-atom segments along the path. The stored reference gives $64.49^{\circ}$ for this path minimum.

\paragraph{Medical Imaging: a named lesion and its spatial extent.}
The medical example supplies the diagnostic target, a small left pleural effusion in a CT volume. It asks for the lesion's extent along the three array axes, every longest axis, and a signed left--right centroid offset. The reference extent is $[86,81,61]$ voxels. Extents include both bounding voxels, whereas the centroid offset is a physical distance in millimetres.

\subsection{Scoring details for related benchmarks}
\label{newapp:related-comparison}
Section~\ref{newsec:related} discusses the scientific question-answering and evidence-access benchmarks. Table~\ref{tab:related-work-comparison} summarizes their scope, objectives, evidence access, and grading.

DeepTumorVQA generates reference answers using deterministic programs applied to validated imaging metadata \citep{Chen2026DeepTumorVQA}. It reports multiple-choice answer accuracy and rule-based scoring of free-form responses.
ProtStructQA supplies reference values through deterministic calculations and checks predictions using answer-type-specific tolerances, exact matching, or set overlap \citep{Mandiga2026ProtStructQA}.

\paragraph{Broader visual question answering.}
ScienceQA combines multimodal science questions with lectures and explanations \citep{lu2022scienceqa}.
MMMU evaluates college-level knowledge across disciplines, while MathVista focuses on mathematical reasoning in visual contexts \citep{yue2024mmmu,lu2024mathvista}.
These benchmarks establish complementary tests of subject knowledge and visual reasoning. ScienceVision instead supplies native scientific 3D artifacts and compares evidence-access settings for fixed questions.

\paragraph{Volumetric inputs and evidence requirements.}
M3D-Bench evaluates eight medical image-analysis tasks, including visual question answering, positioning, and segmentation \citep{bai2024m3d}.
MedFlowBench requires agents to inspect complete studies and submit answers with structured evidence, checked against withheld annotations \citep{Shen2026MedOpenClaw}.
ScienceVision scores requested answer components deterministically; its scores do not independently certify the inspection trajectory or supporting evidence.

\paragraph{Interaction modalities.}
Prior work compares scientific visualization agents using scripts, MCP or API calls, command-line interfaces, and graphical interfaces \citep{Vonderhorst2026SciVisInteraction}.
This makes interaction design an existing evaluation axis. ScienceVision's comparison concerns access to native artifacts, additional viewer tools, and rendered images under shared final-answer checks.

\subsubsection{Question answering and spatial evidence}
\label{newapp:reference-qa}
Classical VQA asks natural-language questions about images, establishing an answer-centred evaluation interface \citep{Antol2015VQA}.
GPQA instead tests expert-written scientific multiple-choice questions in biology, physics, and chemistry \citep{rein2023gpqa}.
Scientists' First Exam separates scientific signal perception, attribute understanding, and comparative reasoning across five disciplines \citep{zhou2025sfe}.
These tasks distinguish scientific knowledge and visual interpretation from inspection of an editable evidence source.
ScienceVision retains predefined questions but supplies native scientific objects alongside the specified evidence-access conditions.

Spatial457 varies object recognition, position, and orientation requirements through seven synthetic question types \citep{wang2025spatial457}.
3DSRBench tests spatial relations in natural images and examines robustness to common and uncommon camera viewpoints \citep{Ma2025ThreeDSRBench}.
MMSI-Bench evaluates questions requiring spatial information across multiple images \citep{Yang2026MMSIBench}.
These benchmarks show that image-based evaluation already probes three-dimensional relations and viewpoint sensitivity.
ScienceVision's distinction is therefore the scientific objects, domain conventions, and access conditions, rather than the presence of spatial reasoning alone.

SLAKE combines medical image questions with physician-annotated semantic labels and a medical knowledge base \citep{liu2021slake}.
3D-RAD extends radiology VQA to CT scans, including medical computation and longitudinal temporal diagnosis \citep{gai20253drad}.
R1Seg-3D aligns dense visual features, textual reasoning, and mask decoding for medical CT segmentation \citep{hao2025r1seg3d}.
Their outputs distinguish medical question answering from dense localization; ScienceVision grades the requested answer fields, without evaluating segmentation quality as a general objective.

\subsubsection{Architecture and mechanical CAD}
\label{newapp:reference-cad}
MMArch asks architecture and civil-engineering questions grounded in figures from peer-reviewed papers \citep{du2026mmarch}.
BIM adaptive exploration instead executes code to discover model structure at runtime and answers questions in the accompanying ifc-bench v2 \citep{Hellin2026BIMAdaptiveExploration}.
Thus, both architectural visual reasoning and question answering over structured building models precede ScienceVision.
Our architecture tasks concern supplied objects and explicit geometric conventions within the same cross-domain scoring protocol.

ABC supplies parametric curves and surfaces for geometric learning, including surface-normal estimation \citep{koch2019abc}.
Fusion 360 Gallery provides human design sequences and a construction environment for recovering CAD programs from target geometry \citep{willis2021fusion}.
These resources establish geometric and construction representations, whereas ScienceVision evaluates questions about supplied CAD artifacts.

VideoCAD studies long-horizon CAD interface interactions and also derives a VQA benchmark for spatial reasoning and video understanding \citep{man2025videocad}.
BenchCAD combines visual and code question answering with image-to-code generation and instruction-guided CAD editing \citep{zhang2026benchcad}.
These benchmarks extend CAD evaluation to question answering as well as construction and editing.
ScienceVision asks questions about fixed supplied artifacts and does not require reconstructing their construction histories.

P3D-Bench evaluates parametric generation from text, images, and assembly specifications through executable, geometric, and structural criteria \citep{yang2026p3dbench}.
CADEngBench tests parametric design and functional editing, finite-element responses, and assembly-joint grounding \citep{Singh2026CADEngBench}.
CADWorld evaluates GUI-based FreeCAD workflows using executable checks on saved projects and auxiliary outputs \citep{Dong2026CADWorld}.
Their engineering outputs extend beyond ScienceVision's final-answer checks; successful artifact creation and successful inspection answer different evaluation questions.

\subsubsection{Scientific structure and generated-object evaluation}
\label{newapp:reference-structure}
QM9 supplies calculated molecular geometries and quantum-chemical properties \citep{ramakrishnan2014qm9}.
ATOM3D organizes learning tasks over three-dimensional biomolecular structures \citep{townshend2021atom3d}.
These resources connect structure to scientific prediction targets, whereas ScienceVision evaluates answers about supplied structures under stated identity and measurement conventions.

SURFACEBENCH recovers symbolic surface equations from sampled three-dimensional data and combines symbolic equivalence with geometric checks \citep{kabra2026surfacebench}.
Its criterion concerns discovering a representation that fits the supplied data; ScienceVision instead checks the requested properties of an existing artifact.
Gen3DEval assesses generated objects for text fidelity, appearance, and surface quality using vision-language models \citep{maiti2025gen3deval}.
Hi3DEval evaluates generated content at object and part levels, including material realism \citep{zhang2025hi3deval}.
These quality assessments differ from checking domain-specific answers against task references, even when both evaluations use rendered views.

\subsubsection{Scientific workflows and evaluation boundaries}
\label{newapp:reference-workflows}
DiscoveryWorld evaluates simulated discovery cycles involving hypotheses, experiments, and conclusions \citep{Jansen2024DiscoveryWorld}.
LabUtopia benchmarks scientific embodied agents from atomic actions through long-horizon laboratory manipulation \citep{Li2025LabUtopia}.
The AI Scientist automates research ideas, code, experiments, and manuscript preparation in machine-learning settings \citep{Lu2024AIScientist}.
Co-Scientist generates and refines hypotheses for scientists, with experimental validation in selected biomedical applications \citep{Gottweis2026CoScientist}.
Foam-Agent automates OpenFOAM simulation workflows and exposes composable functions through MCP \citep{Yue2025FoamAgent}.
ScienceVision's inspection questions do not measure these broader discovery, embodied-control, or simulation-execution objectives.

Execution-based grading also appears in benchmarks outside scientific visualization: SWE-bench evaluates repository issue resolution, while OSWorld checks tasks in real computer environments \citep{jimenez2024swebench,xie2024osworld}.
ScienceVision uses task-specific final-answer checks; deterministic grading alone is therefore not a distinctive contribution relative to agent evaluation generally.
MT-Bench and Chatbot Arena examine model-based judging, including position, verbosity, and self-enhancement biases \citep{Zheng2023Judging}.
Avoiding an LLM judge removes that judging stage, but does not establish the correctness of ScienceVision's references or comparison rules.

Dynabench uses human-and-model interaction to collect examples that challenge a target model \citep{kiela2021dynabench}.
This provides context for interpreting model-informed task selection, although ScienceVision does not implement the Dynabench collection protocol.
Contamination studies distinguish memorization from exploitation and call for benchmark-specific exposure assessment \citep{magar2022contamination,sainz2023contamination}.
ScienceVision's reference provenance is not a training-data exposure audit; the available evidence does not establish that its artifacts or questions were absent from model training.

\subsection{Extended motivation and resource positioning}

Scientific observation begins with a question about an object and the evidence needed to answer it. Identifying atoms hidden in an initial molecular view can determine which bonded path must be analyzed. Locating a lesion in a CT volume requires interpreting anatomy and returning a location in the specified coordinate frame. The task is to connect evidence from the supplied object to a domain-specific conclusion.

For an AI system, this connection can involve selecting a target, revealing a structure, measuring a relation, and applying scientific definitions or conventions. The relevance of a measurement depends on which object was identified and what the question asks. A correct distance between the wrong atoms does not answer a molecular-contact question. Likewise, a plausible anatomical description does not supply a requested voxel location.

Existing evaluations already study scientific structural question answering and agent interaction with native 3D data. DeepTumorVQA evaluates questions about CT volumes, and SciVisAgentBench includes scientific analysis, visualization, and question-answering outcomes across domains \citep{Chen2026DeepTumorVQA,Ai2026SciVisAgentBench}. ScienceVision evaluates scientific questions about supplied 3D objects using checkable answers. It brings five domains into a shared evaluation, allowing comparison across evaluation settings while documenting execution differences.

ScienceVision contains 250 questions across Architecture, Biology, Chemistry, Mechanical CAD, and Medical Imaging, with 50 questions per domain and 22 subdomains. The questions combine structural evidence with domain concepts, object identity, or explicit conventions. Expert-level describes the intended knowledge required to interpret a task; it does not denote measured difficulty relative to a human baseline.

\bench{} evaluates scientific visual question answering (VQA), with predefined questions, response requirements, and reference answers. Models answer the VQA questions using agent harnesses that manage file access, code execution, and tool interactions. In \textbf{Harness + 3D}, models work with native 3D artifacts through the harness's file and code-execution tools. \textbf{Harness + 3D + MCP} additionally exposes viewer-specific tools for changing views, isolating targets, navigating slices, and measuring geometry \citep{ModelContextProtocol2025Spec}. In \textbf{Harness + 2D}, models answer the questions from rendered 2D images using an agent harness. All settings use the same VQA questions, response requirements, and deterministic final-answer scoring rules.

Our contributions are:
\begin{itemize}[leftmargin=*]
    \item \textbf{A cross-disciplinary scientific 3D VQA benchmark.} \bench{} provides 250 expert-level questions across five equally represented domains and 22 subdomains. Each task pairs a native 3D object with a domain-specific question, explicit response requirements, and a structured reference answer. Some tasks connect target identification under occlusion to geometric or topological reasoning.
    \item \textbf{An expert-inspired MCP inspection environment.} Viewer-dependent interfaces expose viewpoint control, target isolation, slice browsing, and scientific measurement. Designed around expert inspection workflows, these interfaces let models select and inspect task-relevant evidence during question answering.
    \item \textbf{A unified protocol for deterministic scientific VQA evaluation.} Task-specific rules verify scored answer components against reference answers and assign partial credit where specified without a model judge. The same questions, response requirements, and scoring rules apply to Harness + 3D, Harness + 3D + MCP, and Harness + 2D. Results are reported by evaluation setting and domain.
\end{itemize}

\section{Construction, References, and Scoring}
\label{newapp:verification}

\subsection{Construction and reference evidence}
\label{newsec:construction}

\subsubsection{Question construction and expert annotation}
We first check that each 3D object supports the intended question and can be inspected in the viewer. Candidate questions come from templates or curated drafts, sometimes with agent assistance. Questions that satisfy the task rules are selected and sent to experts with their supporting objects.

Experts review the targets and domain assumptions, revise the questions where needed, and inspect the 3D evidence to provide reference answers. Inspection records document how they obtained those answers. A subsequent quality review accepts the annotations or returns them for revision.

\subsubsection{Reference answers and recorded inspections}
Each task includes a reference answer, while inspection records show how experts examined the object. Some references are derived through computation or measurement tools; others are recorded directly. These resources support answer checking, but a recorded reference alone does not establish which alternative answers should receive credit.

Of the 250 references, 110 are recorded values, 106 are derived by scripts, and 34 are derived through tools. These categories describe how the references were obtained. They do not establish that every answer can be recomputed from the supplied objects, because not all reference calculations and scoring procedures were re-executed.

Inspection records describe one route to an answer, rather than the minimum work required or the route an agent must follow. Recorded viewer actions also differ from model responses and tool calls. They do not provide a human task-success baseline.

\subsubsection{Verification and accepted answers}
Scoring must account for units, coordinates, and equivalent representations of an answer. A lesion-pointing question may admit several locations, an object may have accepted aliases, and a numerical answer may require a task-specific tolerance. A single reference value does not define all these rules.

The task configurations contain 54 task-specific scoring adjustments, including component rubrics, accepted aliases, and numerical tolerances. These categories overlap. The configurations document intended scoring rules, but do not verify every scoring decision in earlier runs. Section~\ref{newsec:protocol} defines the scoring unit and aggregation.

\subsection{Task-specific scoring and verification scope}
Task scores are defined in Section~\ref{sec:scoring-main}. Scoring rules specify how answers are interpreted, which equivalent labels are accepted, and how components are matched and weighted. Some tasks score only a subset of the recorded answer components. The available documentation does not establish a universal averaging rule or a common rule for excluding responses from scoring. We retain the recorded scores without applying an additional exclusion rule.

\section{Access Interfaces and Fixed Views}
\label{newapp:access}

The inspection interface supports camera control, target selection and isolation, clipping or slice navigation, and scientific measurements. Available operations depend on the viewer and may differ between the expert and agent interfaces. Lists of permitted operations do not verify which tools were available or functioning in every run. Harness + 3D + MCP also permits computation from source data, so its score does not show whether visual inspection supplied the decisive evidence.

\subsection{Evaluation settings}
\textbf{Harness + 3D} supplies the native 3D object to an agent harness that manages file access and code execution, including computation of distances and structural relations. \textbf{Harness + 3D + MCP} adds viewer inspection and measurement tools to this workflow. Tool use is optional, so this condition evaluates a combined workflow that can use either harness-executed analysis or visual inspection.

In \textbf{Harness + 2D}, models receive the question and pre-rendered 2D views through an agent harness. The main comparison uses six nominal views. The camera positions and image order are fixed; agents cannot request another view. Medical tasks pair volume renders with slices, giving twelve images for six nominal views. Appendix~\ref{newapp:access} describes the interfaces and documentation limits. An earlier comparison of aligned and orbit views is analyzed separately because it used a different task set.

\section{Execution Configurations}
\label{newapp:config}
All main results use the same recorded scores, covering 9,000 task--configuration pairs: 250 tasks for each of 12 models under three evaluation settings.

The blog displays the same 36 overall means but rounds task means before computing domain scores. This produces one last-digit difference: Astra Harness + 3D in chemistry is $0.6094$ here and $0.6095$ on the blog. The difference comes from rounding rather than a new evaluation. Reproducing recorded scores does not independently validate the original runs or scoring decisions.

\subsection{Execution settings and incomplete records}
Official model documentation is available for the evaluated Claude Opus 5.5, GPT-6 Astra, and GPT-6.1 Sol \citep{Anthropic2026ClaudeOpus55,OpenAI2026GPT6Astra,OpenAI2026GPT61Sol}.
These system cards describe the model releases and their developer evaluations; they do not establish the configuration, permissions, or model snapshot used in any retained ScienceVision run.

DeepSeek-V4.1-Flash is stored under the legacy key \texttt{deepseek-v4-flash} in the retained score exports. We preserve the key for traceability. The retained records do not identify a fixed model checkpoint.

The study assigns Codex to GPT-6 Astra and GPT-6.1 Sol, and Claude Code to the other ten models under all three conditions. GPT Codex uses maximum reasoning effort; Claude Code effort is not recorded. Harness + 2D records do not identify the harness for every run, so the study-level assignment cannot be independently verified throughout.

Earlier outputs were retained alongside reruns of 39 tasks after question or scoring revisions. The present task set includes 205 original tasks and 45 replacements. Shared task identifiers do not establish that questions, references, and inputs remained identical. The available records cannot fully reconstruct the conditions for every retained run.

The supplementary Fable/Opus Codex runs use high reasoning effort because maximum-effort requests encountered timeouts. Their reasoning budgets therefore differ from those of the GPT Codex runs. We report model responses and completion tokens separately; hidden reasoning-token counts are unavailable.

\subsection{Data for the harness comparison}
\label{newapp:harness-sources}
Table~\ref{tab:ablation-evidence-harness} combines the main GPT/Codex and Claude/Claude Code results with supplementary runs using the alternative harness. The GPT/Claude Code records precede the final 39-task reruns. The Claude/Codex records contain 3,000 trials spanning two models, two access conditions, and three repeats on 250 tasks.

All 16 table cells cover the same 250 task identifiers with three scores per task. We average repeats within each task, then weight tasks equally and retain zero scores. Shared identifiers do not guarantee identical questions, references, inputs, or scoring conditions. High effort is documented for all supplementary Claude/Codex trials; effort remains unknown for the main Claude Code runs.

Proposed revised GPT/Claude Code averages could not be reproduced from the available trial-level corrections. We therefore use the complete per-task records. All displayed means and percentage-point differences are calculated before rounding, without replaying the original runs.

\subsection{Interaction length, tool feedback, and timeouts}
\label{newapp:harness-process}
All 3,000 supplementary Fable/Opus Codex trials include readable trajectories and completion token counts. High reasoning effort is recorded throughout. We count each recorded model message as one response, including the final answer but excluding initialization messages. Tool calls are counted within those responses. Completion tokens measure generated output; the available records do not measure hidden reasoning length.

The 1,243 full-score trials have median values of six model responses and 2,419 completion tokens. The 1,061 zero-score trials have medians of nine responses and 4,628 tokens. The remaining 696 trials receive partial credit. These are descriptive associations across tasks, models, and conditions, rather than evidence that shorter interactions improve answers. There are 47 execution timeouts, each at a 3,600-second limit. These differ from the request timeouts that motivated the high-effort setting. An earlier reasoning-budget pilot lacks recoverable pairing and denominators, so we do not use its reported rates.

One Fable/Codex Harness + 3D trial illustrates recovery from failed data access. After a molecular-analysis library could not be loaded, the agent inspected the atom records and read the data by an alternative method. It selected the largest occupied component excluding water and polymers, then computed the closest atom-pair distance. The answer, $2.229$ angstroms, received full credit under the archived rubric. The sequence contains six model responses and five tool calls. It shows how feedback can support recovery, without establishing a causal benefit from longer interaction or a difference from Claude Code.

The selected pair joins magnesium and histidine. It was not excluded as bonded because the molecular file contained no explicit connectivity records. This reproduces the recorded calculation and score, but does not validate the intended definition of a nonbonded contact.

Codex records combine inspection and computation through general execution tools. The available Claude Code records expose separate reading, editing, and execution operations. This difference concerns how tools are organized. The main Claude Code score records lack matching trajectories, preventing attribution of score differences to operation order or granularity.

\subsection{Models and execution configurations}
\label{newsec:setup}
We evaluate the 12 models in Table~\ref{newtab:main} on 250 tasks under all three evaluation settings, with 50 tasks in each domain. Each configuration combines a model, an agent harness, and an access condition. The harness manages file access, computation, and tool use.

We use Codex for GPT-6 Astra and GPT-6.1 Sol, and Claude Code for the other ten models \citep{OpenAI2026Codex,Anthropic2026ClaudeCode}. The GPT models use maximum reasoning effort; Claude Code effort is not recorded. For Harness + 2D, harness assignments are documented for the study but cannot be verified for every run. Appendix~\ref{newapp:config} gives the available configuration records.

Supplementary runs with alternative harnesses are analyzed separately in Section~\ref{newsec:harness-comparison}.

\section{Complete Results and Sensitivity Analyses}
\label{newapp:full-results}

\subsection{Main scores and execution settings}
\label{newapp:canonical-results}
\begin{figure}[!htbp]
\centering
\includegraphics[width=\linewidth]{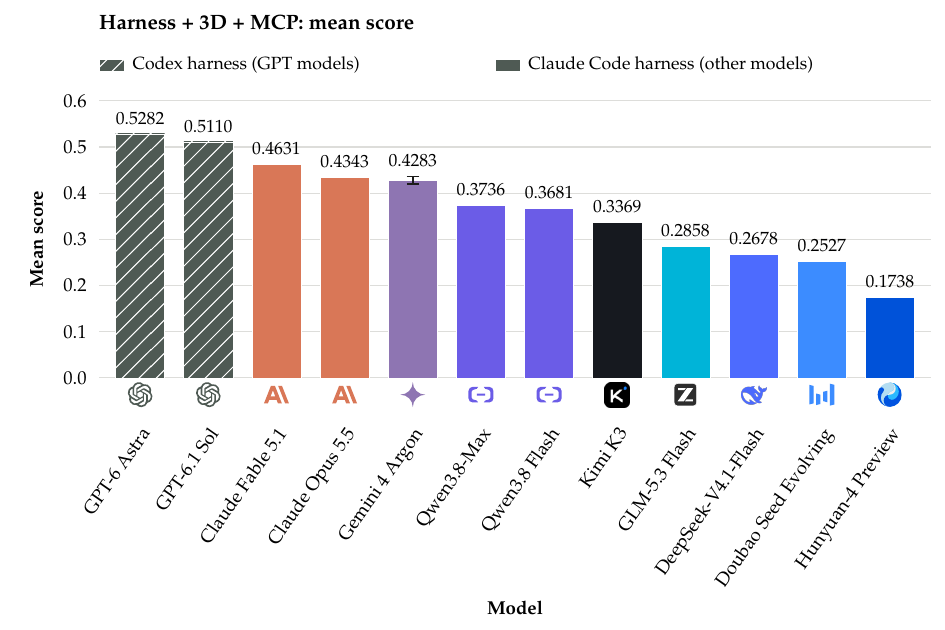}
\caption{\textbf{Harness + 3D + MCP leaderboard.} The 12 canonical model--harness configurations use native 3D artifacts, file access, and code execution with optional MCP inspection and measurement tools. Models are ranked independently by unrounded mean task score; scores match Table~\ref{newtab:main} to four decimals. Whiskers show the min--max of complete-run means over the same 250 tasks. Selected available repeats are averaged within each task, then tasks are weighted equally. Hatched GPT bars use Codex; solid bars use Claude Code. Colors and marks identify providers; the zero-based y-axis matches the other leaderboards. Supplementary cross-harness runs are reported separately. Harness + 3D appears in Figure~\ref{newfig:leaderboard}, and Harness + 2D in Figure~\ref{newfig:fixed6}.}
\label{newfig:leaderboard-mcp}
\end{figure}

\begin{figure}[!htbp]
\centering
\includegraphics[width=\linewidth]{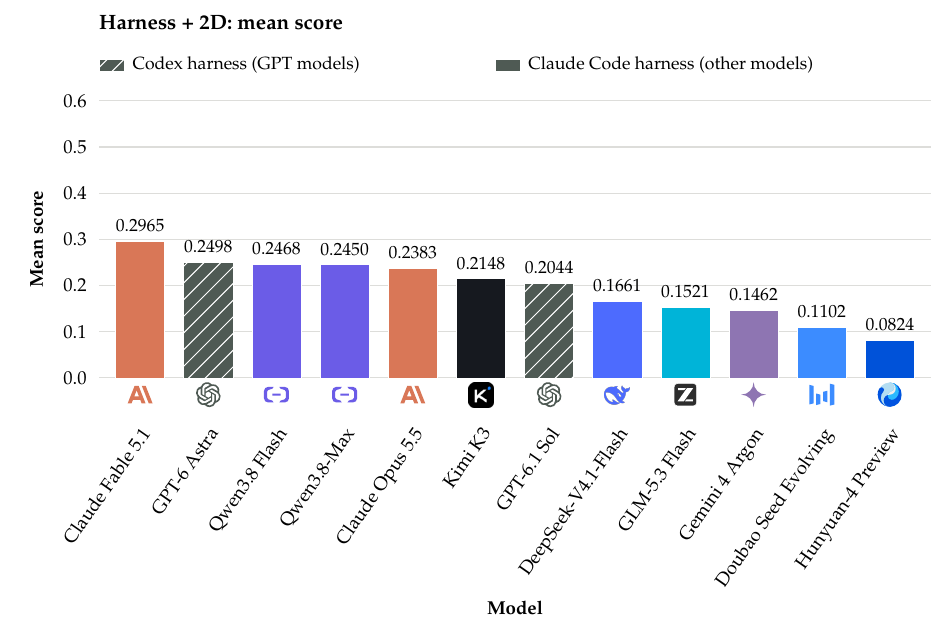}
\caption{\textbf{Harness + 2D leaderboard (mean@3).} The same 12 configurations perform visual question answering from prescribed rendered 2D images using an agent harness. Models are ranked independently by unrounded mean@3; scores match Table~\ref{newtab:main} to four decimals. Under the three-repeat protocol, available scored repeats are averaged within each task, then the 250 tasks are weighted equally. Models lie on the x-axis; bar height is mean@3. Hatched GPT bars use Codex; solid bars use Claude Code. Colors and marks identify providers; the zero-based y-axis matches the other leaderboards. Harness + 3D and Harness + 3D + MCP appear in Figures~\ref{newfig:leaderboard} and~\ref{newfig:leaderboard-mcp}, respectively.}
\label{newfig:fixed6}
\end{figure}

The complete 36 canonical scores are reported in Table~\ref{newtab:main}. The opening Figure~\ref{newfig:leaderboard} ranks Harness + 3D; Figures~\ref{newfig:leaderboard-mcp} and~\ref{newfig:fixed6} below provide independent rankings for Harness + 3D + MCP and Harness + 2D.

\begin{table}[H]
\centering\small
\caption{Mean task scores for 12 models under three access settings on the same 250 tasks. Selected available repeats are averaged within each task, then tasks are weighted equally. Scores are displayed to four decimals. Supplementary runs with alternative harnesses are reported separately.}
\label{newtab:main}
\setlength{\tabcolsep}{0pt}
\begin{tabular*}{\linewidth}{@{\extracolsep{\fill}}lrrr@{}}
\toprule
Model & Harness + 3D & Harness + 3D + MCP & Harness + 2D \\
\midrule
GPT-6 Astra & 0.5291 & 0.5282 & 0.2498 \\
GPT-6.1 Sol & 0.4893 & 0.5110 & 0.2044 \\
Claude Fable 5.1 & 0.4619 & 0.4631 & 0.2965 \\
Claude Opus 5.5 & 0.4539 & 0.4343 & 0.2383 \\
Gemini 4 Argon & 0.4256 & 0.4283 & 0.1462 \\
Qwen3.8-Max & 0.3830 & 0.3736 & 0.2450 \\
Qwen3.8 Flash & 0.3616 & 0.3681 & 0.2468 \\
Kimi K3 & 0.3609 & 0.3369 & 0.2148 \\
GLM-5.3 Flash & 0.2846 & 0.2858 & 0.1521 \\
DeepSeek-V4.1-Flash & 0.2652 & 0.2678 & 0.1661 \\
Doubao Seed Evolving & 0.2435 & 0.2527 & 0.1102 \\
Hunyuan-4 Preview & 0.1598 & 0.1738 & 0.0824 \\
\bottomrule
\end{tabular*}
\end{table}

\subsubsection{Additional context for the canonical scores}
The three leaderboards rank the same 12 selected model--harness configurations separately under each evaluation setting. Appendix~\ref{newapp:canonical-results} gives all three settings and their configurations. Among these canonical leaderboard rows, GPT-6 Astra/Codex has the highest Harness + 3D score, $0.5291$, while Claude Fable~5.1 has the highest Harness + 2D score, $0.2965$. Harness + 3D exceeds Harness + 2D for every model, by $7.73$--$28.49$ percentage points (Table~\ref{newtab:main}).

These mean task scores include partial credit. Harness + 3D permits direct calculation, and execution records are incomplete, so the gaps do not isolate the benefit of visual inspection. Rankings also depend on the execution setting. Supplementary Claude Opus~5.5/Codex Harness + 3D ($0.5448$) and Claude Fable~5.1/Codex Harness + 3D + MCP ($0.5349$) scores exceed the canonical table's maxima. We report those runs separately in Section~\ref{newsec:harness-comparison}.

\paragraph{Access settings.}
Harness + 3D uses an agent harness for file access and computation on native 3D artifacts; Harness + 3D + MCP adds optional inspection and measurement tools. Harness + 2D supplies six fixed rendered views. Medical tasks pair renders and slices, giving twelve images; the available records cannot verify the inputs of every retained run.

\paragraph{Execution settings.}
Table~\ref{newtab:canonical-config} lists the declared harness for each model. The assignments cover all three settings, but Harness + 2D records do not independently verify them for every run. Section~\ref{newsec:protocol} and Appendix~\ref{newapp:config} give execution details and comparison limits.
\begin{table}[H]
\centering\small
\caption{Agent harnesses for each model. Row order matches Table~\ref{newtab:main}. The API ID for Qwen3.8-Max-0902 is \texttt{qwen3.8-max-0902}.}
\label{newtab:canonical-config}
\setlength{\tabcolsep}{0pt}
\begin{tabular*}{\linewidth}{@{\extracolsep{\fill}}ll@{}}
\toprule
Model & Harness \\
\midrule
GPT-6 Astra & Codex (max thinking effort) \\
GPT-6.1 Sol & Codex (max thinking effort) \\
Claude Fable 5.1 & Claude Code \\
Claude Opus 5.5 & Claude Code \\
Gemini 4 Argon & Claude Code \\
Qwen3.8-Max-0902 & Claude Code \\
Qwen3.8 Flash & Claude Code \\
Kimi K3 & Claude Code \\
GLM-5.3 Flash & Claude Code \\
DeepSeek-V4.1-Flash & Claude Code \\
Doubao Seed Evolving & Claude Code \\
Hunyuan-4 Preview & Claude Code \\
\bottomrule
\end{tabular*}
\end{table}

\subsection{Condition-specific domain scores}
\begin{figure}[!htbp]
\centering
\includegraphics[width=\linewidth]{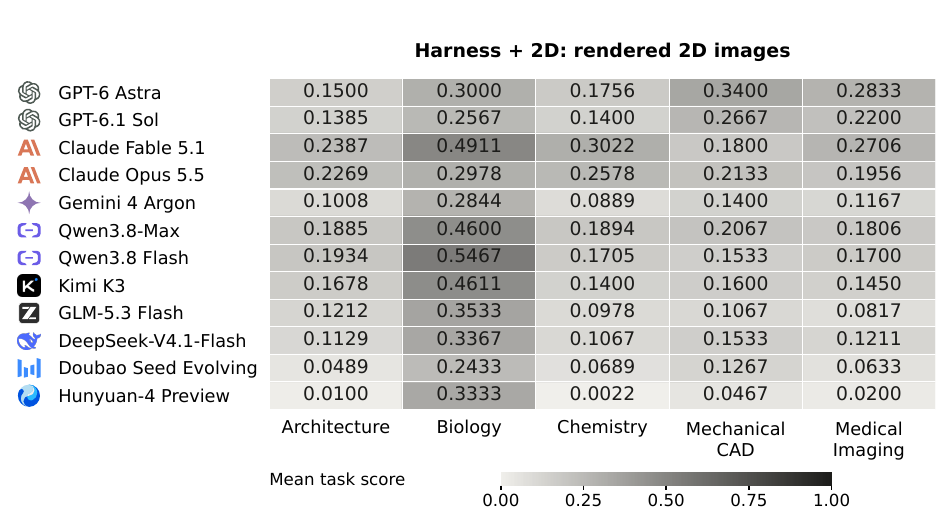}
\caption{Harness + 2D domain mean task scores: rendered 2D images for the 12 canonical model--harness configurations across five domains. Scores are displayed to four decimals on a zero-to-one color scale. Each cell averages 50 equally weighted tasks after averaging the available repeats within each task. Full task-mean precision is retained until display. The score CSV retains all 180 domain means across the three access settings.}
\label{newfig:domains-fixed6}
\end{figure}

Figure~\ref{newfig:domains} shows the score difference after adding MCP for each model and domain. Figure~\ref{newfig:domains-fixed6} gives Harness + 2D domain scores. The domain subtotal rows in Tables~\ref{newtab:subdomains-cc-a}--\ref{newtab:subdomains-cc2d-6-b} retain all 180 absolute means across 12 models, three evaluation settings and five domains. Each domain mean averages 50 tasks after averaging the available repeats within each task. Because domains have equal size, their unrounded means recover the overall score. These values describe the selected questions and tested configurations, rather than population-level differences in disciplinary difficulty.

\subsubsection{Domain variation}
Domain scores show changes that cancel in the overall mean (Figure~\ref{newfig:domains}). With MCP, Astra's Biology score rises from $0.6200$ to $0.6778$, while Medical Imaging falls from $0.3783$ to $0.3406$. Sol's Biology score rises from $0.6000$ to $0.6667$. Among the canonical configurations, Fable~5.1 with MCP has the highest domain mean, $0.6989$ in Biology. Appendix~\ref{newapp:full-results} provides the domain scores for all three conditions.

\subsection{Subdomain Results and Aggregation}
\label{app:subdomain-results}
For these descriptive cross-model comparisons, we average the 12 canonical models equally within each evaluation setting.

\paragraph{Harness + 3D--Harness + 2D gaps depend on the task family.}
The two largest cross-model gaps occur in dihedral measurement ($N=9$; 45.06 percentage points) and conformation/occlusion ($N=21$; 40.68 points). Every model scores higher with Harness + 3D in both subdomains. Biological spatial arrangement ($N=14$) has a much smaller mean gap of 2.58 points, and 5 models score higher with Harness + 2D.

\paragraph{The Harness + 3D + MCP difference changes sign within every model.}
Each of the 12 models has both positive and negative Harness + 3D + MCP minus Harness + 3D differences across the 22 subdomains. Molecular contact ($N=14$) has a mean difference of +8.04 percentage points, with 9 models higher. Dihedral measurement ($N=9$) has a mean difference of $-5.25$ points, with 8 models lower. A single overall MCP difference conceals these opposing local changes.

\paragraph{Lesion pointing remains a shared weakness.}
On the 25 lesion-pointing tasks, mean scores are 0.1000, 0.0978 and 0.0344 for Harness + 3D, Harness + 3D + MCP and Harness + 2D, respectively. All 36 model--condition configurations score lower on lesion pointing than on slice comparison ($N=12$). The Medical Imaging domain mean mixes this localization weakness with higher slice-comparison performance.

\paragraph{Data and aggregation.}
The six tables give 792 mean scores across 22 subdomains, 12 models, and three evaluation settings. They use the same 250 tasks and scores as the main comparison. Each task belongs to one subdomain, as recorded in the accompanying metadata. These groups describe task content rather than mutually exclusive reasoning abilities.

We first average available repeats within each task, as defined in Section~\ref{sec:scoring-main}. For subdomain $u$ with $N_u$ tasks, its mean is $S_u=N_u^{-1}\sum_{i\in u}\bar{s}_i$. Domain and overall means are then
\[
 S_d=\sum_{u\subset d}\frac{N_u}{50}S_u,
 \qquad
 S=\sum_u\frac{N_u}{250}S_u.
\]
Subdomains contain 4 to 25 tasks, so weighting by their task counts is necessary to recover the domain and overall means. Calculations retain full precision until the four-decimal display.

Every model--condition pair covers all 250 tasks, and all subdomain cells are populated. Recorded zero scores are retained. The resulting totals reproduce all 180 domain means and 36 overall means. Appendix~\ref{newapp:config} documents the available execution records. Small subdomain samples and unmatched execution conditions limit generalization beyond this benchmark and prevent causal attribution of access differences.

\begingroup
\small
\setlength{\tabcolsep}{0pt}
\renewcommand{\arraystretch}{1.15}
\setlength{\LTleft}{0pt plus 1fill}
\setlength{\LTright}{0pt plus 1fill}
\setlength{\LTcapwidth}{\textwidth}
\setlength{\LTpre}{6pt}
\setlength{\LTpost}{8pt}
\begin{longtable}{@{}>{\raggedright\arraybackslash}p{0.22\textwidth}>{\raggedleft\arraybackslash}p{0.04\textwidth}*{6}{>{\raggedleft\arraybackslash}p{0.123333\textwidth}}@{}}
\caption{Harness + 3D subdomain mean scores, part 1 of 2. C/O identify the closed/open study groups. Within each subdomain and this evaluation setting, \textbf{bold} and \underline{underlined} scores mark the highest and second-highest distinct four-decimal values within each group across both parts. Ties share emphasis; all-equal groups and summary rows are unmarked. $N$ gives task counts; domain and overall means use $N$ weights.}
\label{newtab:subdomains-cc-a}\label{tab:benchmark-subdomains}\\
\toprule
Subdomain & $N$ & \makecell{GPT-6\\Astra\,(C)} & \makecell{GPT-6.1\\Sol\,(C)} & \makecell{Claude Fable\\5.1\,(C)} & \makecell{Claude Opus\\5.5\,(C)} & \makecell{Gemini 4\\Argon\,(C)} & \makecell{Qwen3.8-\\Max\,(O)} \\
\midrule
\endfirsthead
\caption[]{Harness + 3D subdomain mean scores, part 1 of 2 (continued).}\\
\toprule
Subdomain & $N$ & \makecell{GPT-6\\Astra\,(C)} & \makecell{GPT-6.1\\Sol\,(C)} & \makecell{Claude Fable\\5.1\,(C)} & \makecell{Claude Opus\\5.5\,(C)} & \makecell{Gemini 4\\Argon\,(C)} & \makecell{Qwen3.8-\\Max\,(O)} \\
\midrule
\endhead
\midrule
\multicolumn{8}{r}{\textit{Continued on next page}}\\
\endfoot
\bottomrule
\endlastfoot
\multicolumn{8}{@{}l}{\textbf{Architecture}} \\*
Alignment and stacking & 12 & \underline{0.2639} & 0.2593 & 0.2569 & \textbf{0.3125} & 0.2049 & \textbf{0.3056} \\
Dimensional compliance & 16 & \textbf{0.4071} & 0.3551 & 0.3208 & \underline{0.3781} & 0.3625 & 0.1821 \\
Interference and clearance & 12 & 0.3361 & 0.3222 & \textbf{0.3685} & \underline{0.3569} & 0.3451 & 0.1801 \\
Object geometry readings & 10 & \textbf{0.9167} & 0.8000 & \underline{0.8167} & 0.7333 & 0.5750 & \textbf{0.4833} \\*
\textit{Domain mean} & 50 & 0.4576 & 0.4132 & 0.4161 & 0.4283 & 0.3630 & 0.2715 \\
\addlinespace[3pt]
\multicolumn{8}{@{}l}{\textbf{Biology}} \\*
Interaction typing & 8 & \textbf{0.8750} & \textbf{0.8750} & \underline{0.7500} & \underline{0.7500} & 0.1250 & \textbf{0.7500} \\
Molecular contact & 14 & 0.7857 & 0.7143 & \textbf{0.8333} & \textbf{0.8333} & \underline{0.8214} & \underline{0.7381} \\
Occluded target geometry & 6 & 0.0000 & 0.0000 & \textbf{0.2222} & \underline{0.0926} & 0.0000 & \textbf{0.2778} \\
Proximity choice & 8 & 0.5000 & 0.5000 & \underline{0.5833} & 0.4583 & \textbf{0.6250} & \underline{0.5000} \\
Spatial arrangement & 14 & \textbf{0.6429} & \textbf{0.6429} & \underline{0.6310} & \textbf{0.6429} & 0.6071 & 0.5476 \\*
\textit{Domain mean} & 50 & 0.6200 & 0.6000 & 0.6500 & 0.6178 & 0.5200 & 0.5933 \\
\addlinespace[3pt]
\multicolumn{8}{@{}l}{\textbf{Chemistry}} \\*
Conformation occlusion & 21 & \underline{0.6601} & 0.6283 & 0.5701 & 0.5886 & \textbf{0.6905} & \underline{0.5675} \\
Crystal periodicity & 4 & 0.7222 & \textbf{0.7778} & 0.7222 & 0.7222 & \underline{0.7500} & \underline{0.6111} \\
Dihedral measurement & 9 & \textbf{0.5556} & \textbf{0.5556} & \underline{0.4630} & \textbf{0.5556} & \textbf{0.5556} & \textbf{0.5370} \\
Hybridization state & 7 & \underline{0.8095} & 0.7778 & \textbf{0.9048} & 0.7778 & 0.5476 & 0.4921 \\
Molecular extent & 9 & \underline{0.3395} & \textbf{0.3642} & 0.2778 & 0.2778 & 0.2778 & \underline{0.2778} \\*
\textit{Domain mean} & 50 & 0.6094 & 0.6006 & 0.5572 & 0.5639 & 0.5767 & 0.5028 \\
\addlinespace[3pt]
\multicolumn{8}{@{}l}{\textbf{Mechanical CAD}} \\*
Assembly composition count & 10 & \underline{0.4333} & 0.4000 & 0.3000 & 0.2000 & \textbf{0.5000} & \textbf{0.1667} \\
Mechanism motion & 6 & \textbf{0.5000} & \textbf{0.5000} & \textbf{0.5000} & \underline{0.4444} & 0.2500 & \underline{0.3333} \\
Part instance count & 20 & \textbf{0.5667} & \underline{0.3667} & 0.1667 & 0.0500 & 0.0750 & \underline{0.0667} \\
Part property identification & 14 & \underline{0.7381} & \textbf{0.7619} & 0.6190 & 0.5238 & 0.5714 & \textbf{0.4286} \\*
\textit{Domain mean} & 50 & 0.5800 & 0.5000 & 0.3600 & 0.2600 & 0.3200 & 0.2200 \\
\addlinespace[3pt]
\multicolumn{8}{@{}l}{\textbf{Medical Imaging}} \\*
Anatomy pointing & 5 & \textbf{0.8000} & \textbf{0.8000} & 0.4667 & 0.6667 & \underline{0.7000} & \textbf{0.6667} \\
Geometry measurement & 8 & \textbf{0.5938} & \underline{0.5799} & 0.4965 & 0.5382 & 0.4896 & \textbf{0.5243} \\
Lesion pointing & 25 & \textbf{0.2133} & 0.1467 & 0.1600 & \underline{0.1867} & 0.1200 & \textbf{0.1467} \\
Slice comparison & 12 & 0.4028 & 0.3611 & 0.5000 & \textbf{0.6389} & \underline{0.5833} & 0.4306 \\*
\textit{Domain mean} & 50 & 0.3783 & 0.3328 & 0.3261 & 0.3994 & 0.3483 & 0.3272 \\
\addlinespace[3pt]
\midrule
\textbf{Overall} & 250 & 0.5291 & 0.4893 & 0.4619 & 0.4539 & 0.4256 & 0.3830 \\
\end{longtable}
\endgroup

\begingroup
\small
\setlength{\tabcolsep}{0pt}
\renewcommand{\arraystretch}{1.15}
\setlength{\LTleft}{0pt plus 1fill}
\setlength{\LTright}{0pt plus 1fill}
\setlength{\LTcapwidth}{\textwidth}
\setlength{\LTpre}{6pt}
\setlength{\LTpost}{8pt}
\begin{longtable}{@{}>{\raggedright\arraybackslash}p{0.22\textwidth}>{\raggedleft\arraybackslash}p{0.04\textwidth}*{6}{>{\raggedleft\arraybackslash}p{0.123333\textwidth}}@{}}
\caption{Harness + 3D subdomain mean scores, part 2 of 2. C/O identify the closed/open study groups. Within each subdomain and this evaluation setting, \textbf{bold} and \underline{underlined} scores mark the highest and second-highest distinct four-decimal values within each group across both parts. Ties share emphasis; all-equal groups and summary rows are unmarked. $N$ gives task counts; domain and overall means use $N$ weights.}
\label{newtab:subdomains-cc-b}\\
\toprule
Subdomain & $N$ & \makecell{Qwen3.8\\Flash\,(O)} & \makecell{Kimi\\K3\,(O)} & \makecell{GLM-5.3\\Flash\,(O)} & \makecell{DeepSeek-\\V4.1-\\Flash\,(O)} & \makecell{Doubao Seed\\Evolving\,(O)} & \makecell{Hunyuan-4\\Preview\,(O)} \\
\midrule
\endfirsthead
\caption[]{Harness + 3D subdomain mean scores, part 2 of 2 (continued).}\\
\toprule
Subdomain & $N$ & \makecell{Qwen3.8\\Flash\,(O)} & \makecell{Kimi\\K3\,(O)} & \makecell{GLM-5.3\\Flash\,(O)} & \makecell{DeepSeek-\\V4.1-\\Flash\,(O)} & \makecell{Doubao Seed\\Evolving\,(O)} & \makecell{Hunyuan-4\\Preview\,(O)} \\
\midrule
\endhead
\midrule
\multicolumn{8}{r}{\textit{Continued on next page}}\\
\endfoot
\bottomrule
\endlastfoot
\multicolumn{8}{@{}l}{\textbf{Architecture}} \\*
Alignment and stacking & 12 & \underline{0.2569} & 0.2315 & 0.2116 & 0.1505 & 0.1829 & 0.1319 \\
Dimensional compliance & 16 & \underline{0.2886} & \textbf{0.3179} & 0.2585 & 0.2177 & 0.1836 & 0.0275 \\
Interference and clearance & 12 & \textbf{0.3551} & \underline{0.3074} & 0.2125 & 0.2014 & 0.1343 & 0.1319 \\
Object geometry readings & 10 & \underline{0.2833} & 0.2167 & 0.1667 & 0.2333 & 0.1167 & 0.1167 \\*
\textit{Domain mean} & 50 & 0.2959 & 0.2744 & 0.2178 & 0.2008 & 0.1582 & 0.0955 \\
\addlinespace[3pt]
\multicolumn{8}{@{}l}{\textbf{Biology}} \\*
Interaction typing & 8 & \textbf{0.7500} & \underline{0.6250} & 0.1250 & 0.1250 & 0.0833 & 0.1250 \\
Molecular contact & 14 & \textbf{0.7619} & 0.7143 & \textbf{0.7619} & 0.7143 & 0.6905 & 0.5238 \\
Occluded target geometry & 6 & \underline{0.1296} & 0.0370 & \underline{0.1296} & 0.0926 & 0.0741 & 0.0000 \\
Proximity choice & 8 & 0.4583 & \textbf{0.5417} & 0.4583 & 0.4167 & 0.3333 & 0.3750 \\
Spatial arrangement & 14 & \underline{0.5595} & \textbf{0.5714} & 0.4762 & 0.5476 & 0.4405 & 0.2857 \\*
\textit{Domain mean} & 50 & 0.5789 & 0.5511 & 0.4556 & 0.4511 & 0.3922 & 0.3067 \\
\addlinespace[3pt]
\multicolumn{8}{@{}l}{\textbf{Chemistry}} \\*
Conformation occlusion & 21 & \textbf{0.5781} & 0.5437 & 0.4881 & 0.4008 & 0.2275 & 0.1124 \\
Crystal periodicity & 4 & \textbf{0.6667} & \textbf{0.6667} & 0.4444 & \underline{0.6111} & 0.3333 & 0.2222 \\
Dihedral measurement & 9 & 0.4259 & \underline{0.5000} & 0.3889 & 0.3889 & 0.3889 & 0.4074 \\
Hybridization state & 7 & 0.5238 & 0.5555 & 0.5397 & 0.5238 & \textbf{0.7143} & \underline{0.6667} \\
Molecular extent & 9 & \textbf{0.2963} & \underline{0.2778} & \underline{0.2778} & \underline{0.2778} & 0.1667 & 0.0556 \\*
\textit{Domain mean} & 50 & 0.4995 & 0.4995 & 0.4361 & 0.4106 & 0.3222 & 0.2417 \\
\addlinespace[3pt]
\multicolumn{8}{@{}l}{\textbf{Mechanical CAD}} \\*
Assembly composition count & 10 & 0.0667 & \underline{0.1000} & 0.0333 & \underline{0.1000} & \underline{0.1000} & 0.0000 \\
Mechanism motion & 6 & 0.2222 & \textbf{0.3889} & 0.1111 & 0.1111 & 0.0000 & 0.0000 \\
Part instance count & 20 & 0.0500 & \underline{0.0667} & \textbf{0.1167} & 0.0500 & \underline{0.0667} & 0.0000 \\
Part property identification & 14 & 0.1905 & \underline{0.3333} & 0.1905 & 0.1429 & 0.2619 & 0.2619 \\*
\textit{Domain mean} & 50 & 0.1133 & 0.1867 & 0.1200 & 0.0933 & 0.1200 & 0.0733 \\
\addlinespace[3pt]
\multicolumn{8}{@{}l}{\textbf{Medical Imaging}} \\*
Anatomy pointing & 5 & \textbf{0.6667} & \underline{0.5333} & 0.4667 & 0.1333 & 0.2667 & 0.0000 \\
Geometry measurement & 8 & 0.4410 & 0.4757 & 0.3958 & 0.2500 & \underline{0.5104} & 0.1146 \\
Lesion pointing & 25 & \underline{0.0933} & 0.0400 & 0.0000 & 0.0267 & 0.0400 & 0.0267 \\
Slice comparison & 12 & \underline{0.5694} & \textbf{0.5972} & 0.3472 & 0.4306 & 0.4028 & 0.2083 \\*
\textit{Domain mean} & 50 & 0.3206 & 0.2928 & 0.1933 & 0.1700 & 0.2250 & 0.0817 \\
\addlinespace[3pt]
\midrule
\textbf{Overall} & 250 & 0.3616 & 0.3609 & 0.2846 & 0.2652 & 0.2435 & 0.1598 \\
\end{longtable}
\endgroup

\begingroup
\small
\setlength{\tabcolsep}{0pt}
\renewcommand{\arraystretch}{1.15}
\setlength{\LTleft}{0pt plus 1fill}
\setlength{\LTright}{0pt plus 1fill}
\setlength{\LTcapwidth}{\textwidth}
\setlength{\LTpre}{6pt}
\setlength{\LTpost}{8pt}
\begin{longtable}{@{}>{\raggedright\arraybackslash}p{0.22\textwidth}>{\raggedleft\arraybackslash}p{0.04\textwidth}*{6}{>{\raggedleft\arraybackslash}p{0.123333\textwidth}}@{}}
\caption{Harness + 3D + MCP subdomain mean scores, part 1 of 2. C/O identify the closed/open study groups. Within each subdomain and this evaluation setting, \textbf{bold} and \underline{underlined} scores mark the highest and second-highest distinct four-decimal values within each group across both parts. Ties share emphasis; all-equal groups and summary rows are unmarked. $N$ gives task counts; domain and overall means use $N$ weights.}
\label{newtab:subdomains-ccmcp-a}\\
\toprule
Subdomain & $N$ & \makecell{GPT-6\\Astra\,(C)} & \makecell{GPT-6.1\\Sol\,(C)} & \makecell{Claude Fable\\5.1\,(C)} & \makecell{Claude Opus\\5.5\,(C)} & \makecell{Gemini 4\\Argon\,(C)} & \makecell{Qwen3.8-\\Max\,(O)} \\
\midrule
\endfirsthead
\caption[]{Harness + 3D + MCP subdomain mean scores, part 1 of 2 (continued).}\\
\toprule
Subdomain & $N$ & \makecell{GPT-6\\Astra\,(C)} & \makecell{GPT-6.1\\Sol\,(C)} & \makecell{Claude Fable\\5.1\,(C)} & \makecell{Claude Opus\\5.5\,(C)} & \makecell{Gemini 4\\Argon\,(C)} & \makecell{Qwen3.8-\\Max\,(O)} \\
\midrule
\endhead
\midrule
\multicolumn{8}{r}{\textit{Continued on next page}}\\
\endfoot
\bottomrule
\endlastfoot
\multicolumn{8}{@{}l}{\textbf{Architecture}} \\*
Alignment and stacking & 12 & \underline{0.2708} & 0.2662 & \underline{0.2708} & \textbf{0.3125} & 0.2604 & 0.2245 \\
Dimensional compliance & 16 & \textbf{0.3923} & 0.3539 & 0.3081 & \underline{0.3921} & 0.3625 & 0.2025 \\
Interference and clearance & 12 & 0.3130 & 0.3361 & \textbf{0.3778} & 0.3384 & \underline{0.3535} & 0.2750 \\
Object geometry readings & 10 & \textbf{0.8500} & \underline{0.8167} & \textbf{0.8500} & 0.7167 & 0.5500 & \textbf{0.3167} \\*
\textit{Domain mean} & 50 & 0.4356 & 0.4211 & 0.4243 & 0.4250 & 0.3733 & 0.2480 \\
\addlinespace[3pt]
\multicolumn{8}{@{}l}{\textbf{Biology}} \\*
Interaction typing & 8 & \textbf{0.8750} & \textbf{0.8750} & \underline{0.7500} & 0.7083 & 0.1250 & \textbf{0.7500} \\
Molecular contact & 14 & \textbf{0.9286} & \underline{0.9048} & 0.8810 & 0.8571 & 0.7857 & \textbf{0.8810} \\
Occluded target geometry & 6 & 0.0926 & 0.1111 & \textbf{0.4630} & 0.0926 & \underline{0.2778} & \underline{0.1852} \\
Proximity choice & 8 & \underline{0.5417} & 0.5000 & \textbf{0.6250} & \underline{0.5417} & \textbf{0.6250} & \textbf{0.5417} \\
Spatial arrangement & 14 & \textbf{0.6429} & \textbf{0.6429} & \underline{0.6310} & 0.6071 & 0.6250 & \textbf{0.5476} \\*
\textit{Domain mean} & 50 & 0.6778 & 0.6667 & 0.6989 & 0.6211 & 0.5483 & 0.6289 \\
\addlinespace[3pt]
\multicolumn{8}{@{}l}{\textbf{Chemistry}} \\*
Conformation occlusion & 21 & 0.6230 & \underline{0.6561} & 0.5886 & 0.5556 & \textbf{0.6627} & \textbf{0.4841} \\
Crystal periodicity & 4 & \textbf{0.8333} & \underline{0.8056} & 0.6945 & 0.6945 & 0.7084 & 0.5000 \\
Dihedral measurement & 9 & \underline{0.5000} & \underline{0.5000} & 0.4259 & \underline{0.5000} & \textbf{0.5556} & 0.4074 \\
Hybridization state & 7 & \underline{0.7460} & 0.7143 & 0.7143 & \textbf{0.8413} & 0.5952 & 0.4286 \\
Molecular extent & 9 & \underline{0.3210} & \textbf{0.3395} & 0.2778 & 0.2778 & 0.2778 & \underline{0.2963} \\*
\textit{Domain mean} & 50 & 0.5806 & 0.5911 & 0.5295 & 0.5467 & 0.5683 & 0.4300 \\
\addlinespace[3pt]
\multicolumn{8}{@{}l}{\textbf{Mechanical CAD}} \\*
Assembly composition count & 10 & \textbf{0.4000} & \textbf{0.4000} & \underline{0.3000} & 0.1667 & \textbf{0.4000} & \textbf{0.1000} \\
Mechanism motion & 6 & \textbf{0.5000} & \textbf{0.5000} & \underline{0.4444} & 0.3333 & 0.3333 & \textbf{0.3333} \\
Part instance count & 20 & \textbf{0.6333} & \underline{0.4500} & 0.1500 & 0.0500 & 0.0750 & \underline{0.0833} \\
Part property identification & 14 & \underline{0.7619} & \textbf{0.7857} & 0.6429 & 0.5000 & 0.5357 & \textbf{0.4286} \\*
\textit{Domain mean} & 50 & 0.6067 & 0.5400 & 0.3533 & 0.2333 & 0.3000 & 0.2133 \\
\addlinespace[3pt]
\multicolumn{8}{@{}l}{\textbf{Medical Imaging}} \\*
Anatomy pointing & 5 & \textbf{0.8000} & \underline{0.7333} & 0.5333 & 0.6000 & 0.7000 & \textbf{0.6000} \\
Geometry measurement & 8 & 0.5243 & 0.4965 & \underline{0.5382} & \textbf{0.5556} & 0.4479 & \textbf{0.5903} \\
Lesion pointing & 25 & \underline{0.1600} & \textbf{0.1733} & 0.1333 & 0.1467 & \underline{0.1600} & \underline{0.1333} \\
Slice comparison & 12 & 0.4028 & 0.4028 & 0.4306 & \underline{0.5139} & \textbf{0.5417} & 0.5278 \\*
\textit{Domain mean} & 50 & 0.3406 & 0.3361 & 0.3094 & 0.3456 & 0.3517 & 0.3478 \\
\addlinespace[3pt]
\midrule
\textbf{Overall} & 250 & 0.5282 & 0.5110 & 0.4631 & 0.4343 & 0.4283 & 0.3736 \\
\end{longtable}
\endgroup

\begingroup
\small
\setlength{\tabcolsep}{0pt}
\renewcommand{\arraystretch}{1.15}
\setlength{\LTleft}{0pt plus 1fill}
\setlength{\LTright}{0pt plus 1fill}
\setlength{\LTcapwidth}{\textwidth}
\setlength{\LTpre}{6pt}
\setlength{\LTpost}{8pt}
\begin{longtable}{@{}>{\raggedright\arraybackslash}p{0.22\textwidth}>{\raggedleft\arraybackslash}p{0.04\textwidth}*{6}{>{\raggedleft\arraybackslash}p{0.123333\textwidth}}@{}}
\caption{Harness + 3D + MCP subdomain mean scores, part 2 of 2. C/O identify the closed/open study groups. Within each subdomain and this evaluation setting, \textbf{bold} and \underline{underlined} scores mark the highest and second-highest distinct four-decimal values within each group across both parts. Ties share emphasis; all-equal groups and summary rows are unmarked. $N$ gives task counts; domain and overall means use $N$ weights.}
\label{newtab:subdomains-ccmcp-b}\\
\toprule
Subdomain & $N$ & \makecell{Qwen3.8\\Flash\,(O)} & \makecell{Kimi\\K3\,(O)} & \makecell{GLM-5.3\\Flash\,(O)} & \makecell{DeepSeek-\\V4.1-\\Flash\,(O)} & \makecell{Doubao Seed\\Evolving\,(O)} & \makecell{Hunyuan-4\\Preview\,(O)} \\
\midrule
\endfirsthead
\caption[]{Harness + 3D + MCP subdomain mean scores, part 2 of 2 (continued).}\\
\toprule
Subdomain & $N$ & \makecell{Qwen3.8\\Flash\,(O)} & \makecell{Kimi\\K3\,(O)} & \makecell{GLM-5.3\\Flash\,(O)} & \makecell{DeepSeek-\\V4.1-\\Flash\,(O)} & \makecell{Doubao Seed\\Evolving\,(O)} & \makecell{Hunyuan-4\\Preview\,(O)} \\
\midrule
\endhead
\midrule
\multicolumn{8}{r}{\textit{Continued on next page}}\\
\endfoot
\bottomrule
\endlastfoot
\multicolumn{8}{@{}l}{\textbf{Architecture}} \\*
Alignment and stacking & 12 & \textbf{0.2500} & \underline{0.2315} & 0.2245 & 0.2060 & 0.2245 & 0.1134 \\
Dimensional compliance & 16 & \underline{0.3445} & \textbf{0.3455} & 0.2512 & 0.1684 & 0.2175 & 0.1033 \\
Interference and clearance & 12 & \textbf{0.3778} & \underline{0.3454} & 0.1977 & 0.2000 & 0.1181 & 0.0556 \\
Object geometry readings & 10 & \textbf{0.3167} & \underline{0.2667} & 0.2000 & 0.0667 & 0.1167 & 0.0667 \\*
\textit{Domain mean} & 50 & 0.3242 & 0.3023 & 0.2217 & 0.1647 & 0.1751 & 0.0870 \\
\addlinespace[3pt]
\multicolumn{8}{@{}l}{\textbf{Biology}} \\*
Interaction typing & 8 & \underline{0.7083} & 0.5417 & 0.1250 & 0.1250 & 0.0833 & 0.1250 \\
Molecular contact & 14 & \underline{0.8571} & 0.8333 & 0.6905 & 0.7143 & 0.7619 & 0.7619 \\
Occluded target geometry & 6 & 0.1296 & \underline{0.1852} & \textbf{0.2037} & 0.0741 & 0.0926 & 0.1481 \\
Proximity choice & 8 & \underline{0.5000} & \textbf{0.5417} & 0.4583 & \underline{0.5000} & 0.2083 & 0.4167 \\
Spatial arrangement & 14 & \textbf{0.5476} & 0.5119 & \underline{0.5357} & 0.5119 & 0.4286 & 0.3214 \\*
\textit{Domain mean} & 50 & 0.6022 & 0.5722 & 0.4611 & 0.4522 & 0.3911 & 0.4078 \\
\addlinespace[3pt]
\multicolumn{8}{@{}l}{\textbf{Chemistry}} \\*
Conformation occlusion & 21 & \underline{0.4815} & 0.4709 & 0.4312 & 0.3585 & 0.3585 & 0.2050 \\
Crystal periodicity & 4 & \textbf{0.6667} & \underline{0.5556} & \textbf{0.6667} & \textbf{0.6667} & 0.3889 & 0.1111 \\
Dihedral measurement & 9 & 0.3148 & 0.3148 & \textbf{0.4630} & 0.3889 & 0.2778 & \underline{0.4444} \\
Hybridization state & 7 & 0.5238 & 0.5397 & \underline{0.6508} & 0.5873 & \textbf{0.6825} & 0.4762 \\
Molecular extent & 9 & \underline{0.2963} & \textbf{0.3148} & 0.2222 & 0.2778 & 0.1852 & 0.1481 \\*
\textit{Domain mean} & 50 & 0.4389 & 0.4311 & 0.4489 & 0.4061 & 0.3606 & 0.2683 \\
\addlinespace[3pt]
\multicolumn{8}{@{}l}{\textbf{Mechanical CAD}} \\*
Assembly composition count & 10 & \underline{0.0667} & 0.0333 & \underline{0.0667} & \underline{0.0667} & 0.0000 & 0.0333 \\
Mechanism motion & 6 & \underline{0.2222} & \underline{0.2222} & 0.0000 & 0.0556 & 0.1111 & 0.0000 \\
Part instance count & 20 & \textbf{0.1000} & 0.0500 & \underline{0.0833} & \underline{0.0833} & 0.0500 & 0.0000 \\
Part property identification & 14 & 0.1667 & 0.3333 & 0.0714 & 0.2619 & \underline{0.3571} & 0.1190 \\*
\textit{Domain mean} & 50 & 0.1267 & 0.1467 & 0.0667 & 0.1267 & 0.1333 & 0.0400 \\
\addlinespace[3pt]
\multicolumn{8}{@{}l}{\textbf{Medical Imaging}} \\*
Anatomy pointing & 5 & \underline{0.5333} & 0.3333 & 0.3333 & 0.1333 & 0.4667 & 0.0000 \\
Geometry measurement & 8 & \underline{0.4688} & 0.4306 & 0.3785 & 0.1840 & 0.3750 & 0.1424 \\
Lesion pointing & 25 & \textbf{0.1600} & 0.0267 & 0.0400 & 0.0267 & 0.0133 & 0.0000 \\
Slice comparison & 12 & \textbf{0.5833} & 0.4861 & 0.4861 & \underline{0.5556} & 0.3750 & 0.1806 \\*
\textit{Domain mean} & 50 & 0.3483 & 0.2322 & 0.2306 & 0.1894 & 0.2033 & 0.0661 \\
\addlinespace[3pt]
\midrule
\textbf{Overall} & 250 & 0.3681 & 0.3369 & 0.2858 & 0.2678 & 0.2527 & 0.1738 \\
\end{longtable}
\endgroup

\begingroup
\small
\setlength{\tabcolsep}{0pt}
\renewcommand{\arraystretch}{1.15}
\setlength{\LTleft}{0pt plus 1fill}
\setlength{\LTright}{0pt plus 1fill}
\setlength{\LTcapwidth}{\textwidth}
\setlength{\LTpre}{6pt}
\setlength{\LTpost}{8pt}
\begin{longtable}{@{}>{\raggedright\arraybackslash}p{0.22\textwidth}>{\raggedleft\arraybackslash}p{0.04\textwidth}*{6}{>{\raggedleft\arraybackslash}p{0.123333\textwidth}}@{}}
\caption{Harness + 2D subdomain mean scores, part 1 of 2. C/O identify the closed/open study groups. Within each subdomain and this evaluation setting, \textbf{bold} and \underline{underlined} scores mark the highest and second-highest distinct four-decimal values within each group across both parts. Ties share emphasis; all-equal groups and summary rows are unmarked. $N$ gives task counts; domain and overall means use $N$ weights.}
\label{newtab:subdomains-cc2d-6-a}\\
\toprule
Subdomain & $N$ & \makecell{GPT-6\\Astra\,(C)} & \makecell{GPT-6.1\\Sol\,(C)} & \makecell{Claude Fable\\5.1\,(C)} & \makecell{Claude Opus\\5.5\,(C)} & \makecell{Gemini 4\\Argon\,(C)} & \makecell{Qwen3.8-\\Max\,(O)} \\
\midrule
\endfirsthead
\caption[]{Harness + 2D subdomain mean scores, part 1 of 2 (continued).}\\
\toprule
Subdomain & $N$ & \makecell{GPT-6\\Astra\,(C)} & \makecell{GPT-6.1\\Sol\,(C)} & \makecell{Claude Fable\\5.1\,(C)} & \makecell{Claude Opus\\5.5\,(C)} & \makecell{Gemini 4\\Argon\,(C)} & \makecell{Qwen3.8-\\Max\,(O)} \\
\midrule
\endhead
\midrule
\multicolumn{8}{r}{\textit{Continued on next page}}\\
\endfoot
\bottomrule
\endlastfoot
\multicolumn{8}{@{}l}{\textbf{Architecture}} \\*
Alignment and stacking & 12 & 0.1528 & 0.1361 & \underline{0.1815} & \textbf{0.1894} & 0.0972 & \textbf{0.1931} \\
Dimensional compliance & 16 & 0.1251 & 0.1085 & \textbf{0.2211} & \underline{0.1851} & 0.0875 & 0.1172 \\
Interference and clearance & 12 & 0.1806 & 0.1713 & \underline{0.2963} & \textbf{0.3009} & 0.1644 & 0.2417 \\
Object geometry readings & 10 & 0.1500 & 0.1500 & \textbf{0.2667} & \underline{0.2500} & 0.0500 & \textbf{0.2333} \\*
\textit{Domain mean} & 50 & 0.1500 & 0.1385 & 0.2387 & 0.2269 & 0.1008 & 0.1885 \\
\addlinespace[3pt]
\multicolumn{8}{@{}l}{\textbf{Biology}} \\*
Interaction typing & 8 & \underline{0.3750} & 0.2917 & \textbf{0.7500} & \underline{0.3750} & 0.1250 & \underline{0.6667} \\
Molecular contact & 14 & 0.0000 & 0.0000 & \textbf{0.4524} & 0.0000 & \underline{0.0952} & 0.5000 \\
Occluded target geometry & 6 & 0.0000 & 0.0000 & \textbf{0.3148} & 0.0926 & \underline{0.1481} & \underline{0.1667} \\
Proximity choice & 8 & \underline{0.4583} & 0.4167 & 0.2500 & 0.4167 & \textbf{0.5833} & \textbf{0.4583} \\
Spatial arrangement & 14 & \textbf{0.5952} & 0.5119 & \textbf{0.5952} & \underline{0.5714} & 0.4524 & 0.4286 \\*
\textit{Domain mean} & 50 & 0.3000 & 0.2567 & 0.4911 & 0.2978 & 0.2844 & 0.4600 \\
\addlinespace[3pt]
\multicolumn{8}{@{}l}{\textbf{Chemistry}} \\*
Conformation occlusion & 21 & 0.1058 & 0.0794 & \textbf{0.2883} & \underline{0.2010} & 0.0529 & \textbf{0.1468} \\
Crystal periodicity & 4 & 0.0000 & 0.0000 & \textbf{0.5000} & \textbf{0.5000} & \underline{0.3889} & \textbf{0.4722} \\
Dihedral measurement & 9 & \textbf{0.0741} & \textbf{0.0741} & \underline{0.0370} & 0.0000 & 0.0000 & \underline{0.0370} \\
Hybridization state & 7 & \textbf{0.8413} & 0.6667 & 0.7460 & \underline{0.8095} & 0.2540 & \textbf{0.5238} \\
Molecular extent & 9 & 0.0000 & 0.0000 & \textbf{0.1667} & \underline{0.1111} & 0.0000 & \textbf{0.0556} \\*
\textit{Domain mean} & 50 & 0.1756 & 0.1400 & 0.3022 & 0.2578 & 0.0889 & 0.1894 \\
\addlinespace[3pt]
\multicolumn{8}{@{}l}{\textbf{Mechanical CAD}} \\*
Assembly composition count & 10 & \textbf{0.4333} & \underline{0.3000} & \underline{0.3000} & 0.2667 & 0.1333 & \underline{0.3333} \\
Mechanism motion & 6 & \underline{0.4444} & 0.3333 & \textbf{0.5000} & 0.3333 & 0.2778 & \underline{0.2778} \\
Part instance count & 20 & \textbf{0.0500} & \textbf{0.0500} & \underline{0.0000} & \underline{0.0000} & \underline{0.0000} & \underline{0.0000} \\
Part property identification & 14 & \textbf{0.6429} & \underline{0.5238} & 0.2143 & 0.4286 & 0.2857 & \textbf{0.3810} \\*
\textit{Domain mean} & 50 & 0.3400 & 0.2667 & 0.1800 & 0.2133 & 0.1400 & 0.2067 \\
\addlinespace[3pt]
\multicolumn{8}{@{}l}{\textbf{Medical Imaging}} \\*
Anatomy pointing & 5 & \underline{0.6667} & \textbf{0.7333} & \textbf{0.7333} & 0.5333 & 0.0000 & \textbf{0.1333} \\
Geometry measurement & 8 & 0.0417 & 0.0417 & \textbf{0.2535} & \underline{0.1389} & 0.1250 & \textbf{0.2326} \\
Lesion pointing & 25 & \textbf{0.2000} & \underline{0.1200} & 0.0533 & 0.0267 & 0.0000 & \underline{0.0000} \\
Slice comparison & 12 & \underline{0.4583} & 0.3333 & \textbf{0.5417} & 0.4444 & 0.4028 & \underline{0.5417} \\*
\textit{Domain mean} & 50 & 0.2833 & 0.2200 & 0.2706 & 0.1956 & 0.1167 & 0.1806 \\
\addlinespace[3pt]
\midrule
\textbf{Overall} & 250 & 0.2498 & 0.2044 & 0.2965 & 0.2383 & 0.1462 & 0.2450 \\
\end{longtable}
\endgroup

\begingroup
\small
\setlength{\tabcolsep}{0pt}
\renewcommand{\arraystretch}{1.15}
\setlength{\LTleft}{0pt plus 1fill}
\setlength{\LTright}{0pt plus 1fill}
\setlength{\LTcapwidth}{\textwidth}
\setlength{\LTpre}{6pt}
\setlength{\LTpost}{8pt}
\begin{longtable}{@{}>{\raggedright\arraybackslash}p{0.22\textwidth}>{\raggedleft\arraybackslash}p{0.04\textwidth}*{6}{>{\raggedleft\arraybackslash}p{0.123333\textwidth}}@{}}
\caption{Harness + 2D subdomain mean scores, part 2 of 2. C/O identify the closed/open study groups. Within each subdomain and this evaluation setting, \textbf{bold} and \underline{underlined} scores mark the highest and second-highest distinct four-decimal values within each group across both parts. Ties share emphasis; all-equal groups and summary rows are unmarked. $N$ gives task counts; domain and overall means use $N$ weights.}
\label{newtab:subdomains-cc2d-6-b}\\
\toprule
Subdomain & $N$ & \makecell{Qwen3.8\\Flash\,(O)} & \makecell{Kimi\\K3\,(O)} & \makecell{GLM-5.3\\Flash\,(O)} & \makecell{DeepSeek-\\V4.1-\\Flash\,(O)} & \makecell{Doubao Seed\\Evolving\,(O)} & \makecell{Hunyuan-4\\Preview\,(O)} \\
\midrule
\endfirsthead
\caption[]{Harness + 2D subdomain mean scores, part 2 of 2 (continued).}\\
\toprule
Subdomain & $N$ & \makecell{Qwen3.8\\Flash\,(O)} & \makecell{Kimi\\K3\,(O)} & \makecell{GLM-5.3\\Flash\,(O)} & \makecell{DeepSeek-\\V4.1-\\Flash\,(O)} & \makecell{Doubao Seed\\Evolving\,(O)} & \makecell{Hunyuan-4\\Preview\,(O)} \\
\midrule
\endhead
\midrule
\multicolumn{8}{r}{\textit{Continued on next page}}\\
\endfoot
\bottomrule
\endlastfoot
\multicolumn{8}{@{}l}{\textbf{Architecture}} \\*
Alignment and stacking & 12 & \underline{0.1486} & 0.1259 & 0.0810 & 0.1343 & 0.0694 & 0.0139 \\
Dimensional compliance & 16 & \textbf{0.1680} & \underline{0.1600} & 0.0568 & 0.0925 & 0.0104 & 0.0000 \\
Interference and clearance & 12 & \textbf{0.3083} & \underline{0.2625} & 0.1954 & 0.1574 & 0.1065 & 0.0278 \\
Object geometry readings & 10 & 0.1500 & 0.1167 & \underline{0.1833} & 0.0667 & 0.0167 & 0.0000 \\*
\textit{Domain mean} & 50 & 0.1934 & 0.1678 & 0.1212 & 0.1129 & 0.0489 & 0.0100 \\
\addlinespace[3pt]
\multicolumn{8}{@{}l}{\textbf{Biology}} \\*
Interaction typing & 8 & \textbf{0.7083} & 0.3750 & 0.0417 & 0.0833 & 0.0000 & 0.0833 \\
Molecular contact & 14 & \textbf{0.6429} & \underline{0.5952} & 0.5000 & 0.4524 & 0.3571 & 0.5714 \\
Occluded target geometry & 6 & \textbf{0.2222} & 0.0926 & 0.0556 & 0.0000 & 0.0000 & 0.0000 \\
Proximity choice & 8 & \underline{0.4167} & \textbf{0.4583} & 0.2500 & 0.2500 & 0.0833 & 0.2500 \\
Spatial arrangement & 14 & \textbf{0.5714} & 0.5357 & \textbf{0.5714} & \underline{0.5595} & 0.4643 & 0.4286 \\*
\textit{Domain mean} & 50 & 0.5467 & 0.4611 & 0.3533 & 0.3367 & 0.2433 & 0.3333 \\
\addlinespace[3pt]
\multicolumn{8}{@{}l}{\textbf{Chemistry}} \\*
Conformation occlusion & 21 & \textbf{0.1468} & 0.0529 & 0.0370 & \underline{0.0635} & 0.0000 & 0.0000 \\
Crystal periodicity & 4 & 0.4166 & \underline{0.4167} & 0.1944 & 0.1667 & 0.0556 & 0.0000 \\
Dihedral measurement & 9 & 0.0185 & \textbf{0.0741} & 0.0000 & 0.0000 & 0.0000 & 0.0000 \\
Hybridization state & 7 & 0.4921 & \underline{0.5079} & 0.4762 & 0.4762 & 0.4603 & 0.0159 \\
Molecular extent & 9 & \underline{0.0185} & 0.0000 & 0.0000 & 0.0000 & 0.0000 & 0.0000 \\*
\textit{Domain mean} & 50 & 0.1705 & 0.1400 & 0.0978 & 0.1067 & 0.0689 & 0.0022 \\
\addlinespace[3pt]
\multicolumn{8}{@{}l}{\textbf{Mechanical CAD}} \\*
Assembly composition count & 10 & 0.3000 & 0.1333 & 0.1667 & \textbf{0.3667} & 0.1333 & 0.0000 \\
Mechanism motion & 6 & \underline{0.2778} & \textbf{0.4444} & 0.0000 & 0.1111 & 0.1111 & 0.0000 \\
Part instance count & 20 & \underline{0.0000} & \underline{0.0000} & \underline{0.0000} & \textbf{0.0167} & \underline{0.0000} & \underline{0.0000} \\
Part property identification & 14 & 0.2143 & 0.2857 & 0.2619 & 0.2143 & \underline{0.3095} & 0.1667 \\*
\textit{Domain mean} & 50 & 0.1533 & 0.1600 & 0.1067 & 0.1533 & 0.1267 & 0.0467 \\
\addlinespace[3pt]
\multicolumn{8}{@{}l}{\textbf{Medical Imaging}} \\*
Anatomy pointing & 5 & \textbf{0.1333} & \underline{0.0000} & \underline{0.0000} & \underline{0.0000} & \underline{0.0000} & \underline{0.0000} \\
Geometry measurement & 8 & 0.1458 & \underline{0.1562} & 0.0521 & 0.0694 & 0.0000 & 0.0000 \\
Lesion pointing & 25 & \underline{0.0000} & \textbf{0.0133} & \underline{0.0000} & \underline{0.0000} & \underline{0.0000} & \underline{0.0000} \\
Slice comparison & 12 & \textbf{0.5556} & 0.4722 & 0.3056 & 0.4583 & 0.2639 & 0.0833 \\*
\textit{Domain mean} & 50 & 0.1700 & 0.1450 & 0.0817 & 0.1211 & 0.0633 & 0.0200 \\
\addlinespace[3pt]
\midrule
\textbf{Overall} & 250 & 0.2468 & 0.2148 & 0.1521 & 0.1661 & 0.1102 & 0.0824 \\
\end{longtable}
\endgroup

\subsection{Task score profile}
For each task, we take the highest mean across the 36 main configurations. Table~\ref{newtab:profile} groups tasks by whether that maximum is zero, between zero and one, or one. The counts are 34, 83, and 133. A maximum of one requires a full-score mean for at least one configuration; a single successful repeat is insufficient.

\begin{table}[t]
\centering\small
\caption{Descriptive score profile across the 36 canonical configurations. Each task is assigned according to its highest available-repeat mean: zero, strictly between zero and one, or one. A full mean for one configuration does not imply success for every model or individual repeat.}
\label{newtab:profile}
\setlength{\tabcolsep}{0pt}
\begin{tabular*}{\linewidth}{@{\extracolsep{\fill}}lrrr@{}}
\toprule
Domain & All zero & Partial maximum & Full-score maximum \\
\midrule
Architecture & 6 & 27 & 17 \\
Biology & 4 & 11 & 35 \\
Chemistry & 6 & 21 & 23 \\
Mechanical CAD & 5 & 10 & 35 \\
Medical Imaging & 13 & 14 & 23 \\
\bottomrule
\end{tabular*}
\end{table}

\begin{figure}[t]
\centering
\includegraphics[width=\linewidth]{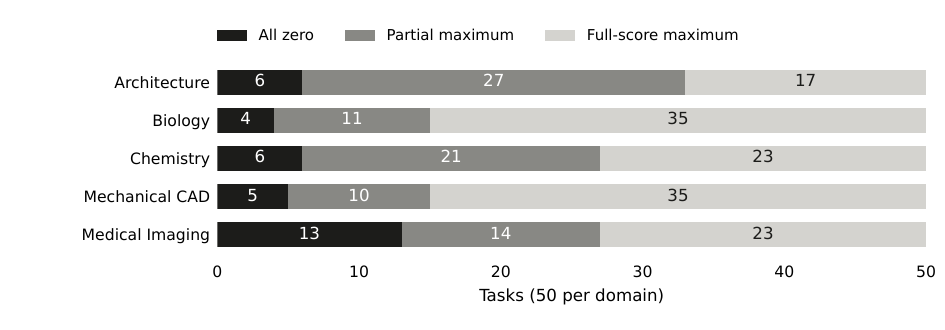}
\caption{Task score profiles within each domain. Every task is assigned by its highest available-repeat mean across the 36 canonical configurations: zero (dark), strictly between zero and one (mid-grey), or one (light). Segment labels count tasks, with 50 tasks per row. These are descriptive categories for the evaluated cohort. Table~\ref{newtab:profile} gives the counts.}
\label{newfig:profile}
\end{figure}

Across the 36 configurations, 34 tasks receive a mean score of zero everywhere. At least one configuration reaches a task mean of one on 133 tasks. The remaining 83 tasks have a best mean between zero and one. Medical Imaging accounts for 13 all-zero tasks, compared with four to six in each other domain (Figure~\ref{newfig:profile}). These counts characterize the selected tasks and tested systems, rather than intrinsic domain difficulty.

Reference inspections show how an expert approached a question, but cannot establish what an evaluated model observed. Answer scores alone also cannot distinguish failures to acquire evidence from failures of scientific inference.

\subsection{CAD target-definition sensitivity}
The prior audit examined eight CAD questions with unclear targets or counting units. Seven stored counts matched the number of solid bodies across the assembled object. For the remaining question, the reference of seven counted distinct immediate component definitions, whereas the whole assembly contained 26 solid bodies. Without an explicit target and counting unit, reproducing a stored number does not resolve the intended answer. The audit therefore did not establish a numerical reference error.

The diagnostic excludes these eight questions and weights the remaining 242 tasks equally: 42 CAD tasks and 50 in each other domain. It preserves all 12-model rankings, every Harness + 3D--Harness + 2D direction, and all MCP difference signs. The largest mean change is $1.531$ percentage points for Fable Harness + 3D + MCP. The benchmark questions, references, and reported 250-task scores remain unchanged.

\begin{table}[t]
\centering\small
\caption{Published 250-task means and the 242-task diagnostic excluding eight CAD questions with underspecified targets or counting units. The diagnostic weights 42 CAD and 50 tasks from each other domain equally per task. The exclusion neither changes the published benchmark nor certifies the remaining references.}
\label{newtab:cad8}
\setlength{\tabcolsep}{0pt}
\begin{tabular*}{\linewidth}{@{\extracolsep{\fill}}lrrrrrr@{}}
\toprule
 & \multicolumn{2}{c}{Harness + 3D} & \multicolumn{2}{c}{Harness + 3D + MCP} & \multicolumn{2}{c}{Harness + 2D} \\ Model & 250 & 242 & 250 & 242 & 250 & 242 \\
\midrule
GPT-6 Astra & 0.5291 & 0.5356 & 0.5282 & 0.5292 & 0.2498 & 0.2580 \\
GPT-6.1 Sol & 0.4893 & 0.5013 & 0.5110 & 0.5183 & 0.2044 & 0.2097 \\
Claude Fable 5.1 & 0.4619 & 0.4772 & 0.4631 & 0.4784 & 0.2965 & 0.3063 \\
Claude Opus 5.5 & 0.4539 & 0.4689 & 0.4343 & 0.4487 & 0.2383 & 0.2461 \\
Gemini 4 Argon & 0.4256 & 0.4397 & 0.4283 & 0.4425 & 0.1462 & 0.1510 \\
Qwen3.8-Max & 0.3830 & 0.3956 & 0.3736 & 0.3860 & 0.2450 & 0.2531 \\
Qwen3.8 Flash & 0.3616 & 0.3736 & 0.3681 & 0.3802 & 0.2468 & 0.2550 \\
Kimi K3 & 0.3609 & 0.3728 & 0.3369 & 0.3481 & 0.2148 & 0.2219 \\
GLM-5.3 Flash & 0.2846 & 0.2940 & 0.2858 & 0.2952 & 0.1521 & 0.1572 \\
DeepSeek-V4.1-Flash & 0.2652 & 0.2739 & 0.2678 & 0.2767 & 0.1661 & 0.1703 \\
Doubao Seed Evolving & 0.2435 & 0.2516 & 0.2527 & 0.2610 & 0.1102 & 0.1139 \\
Hunyuan-4 Preview & 0.1598 & 0.1650 & 0.1738 & 0.1796 & 0.0824 & 0.0852 \\
\bottomrule
\end{tabular*}
\end{table}

Supporting materials identify the eight questions and document the independent component counts. Recomputing this score diagnostic uses the recorded scores; repeating the structural audit additionally requires the original CAD objects.

\subsection{Scope of the execution-configuration comparison}
The harness comparison in Section~\ref{newsec:harness-comparison} averages three repeats per task and weights all 250 tasks equally. We report the observed differences without a paired significance test or a causal estimate of harness effects.

\subsection{Additional protocol comparison details}

Harness + 3D and Harness + 2D differ in both evidence representation and permitted computation. Run records also leave uncertainty about whether execution conditions, inputs, and scoring were fully matched. We therefore interpret score differences as comparisons of evaluated configurations. They do not isolate effects of active observation, MCP access, or the model alone.

\subsubsection{Additional score comparisons}
A prior audit of eight CAD questions found underspecified targets or counting units, without establishing numerical errors in their reference answers. Excluding those questions leaves 242 tasks and preserves all three model rankings, every Harness + 3D--Harness + 2D direction, and the eight-increase/four-decrease MCP split.

Astra's Harness + 3D score changes from $0.5291$ to $0.5356$; Fable's Harness + 2D score changes from $0.2965$ to $0.3063$. Appendix~\ref{newapp:full-results} gives the complete diagnostic. This check leaves the benchmark unchanged and does not validate all remaining references.

\section{Execution Configurations and Earlier Design Studies}
\label{newapp:historical}

\label{newsec:ablations}
We examine execution settings and two earlier studies of view budgets and tool-only access. The earlier studies used different task sets and cannot estimate matched effects on the present benchmark.

\subsection{Execution harness and reasoning budget}
\label{newsec:harness-comparison}
The alternative-harness runs cover Harness + 3D and Harness + 3D + MCP for four models (Table~\ref{tab:ablation-evidence-harness}). GPT models use maximum reasoning effort in Codex. The supplementary Fable and Opus runs use high effort because maximum-effort requests encountered timeouts. The main leaderboard retains one declared harness per model, with these supplementary runs reported separately.

Codex scores higher in all eight model--access comparisons, by $4.56$--$13.40$ percentage points. Supplementary Opus/Codex Harness + 3D reaches $0.5448$, exceeding the main Harness + 3D maximum of $0.5291$. Differences in execution settings and evaluation records prevent attribution to a controlled harness effect. Appendix~\ref{newapp:harness-sources} describes the comparison data.

Additional interaction sometimes helps an agent recover from an error. In one Fable/Codex trial, a failed attempt to read molecular data led to further inspection and an alternative distance calculation. The answer received full credit under the recorded rubric. Yet among 3,000 supplementary Fable/Opus Codex trials, full-score trials had a median of six model responses, compared with nine for zero-score trials. In that batch, 47 trials ended in timeouts. These associations do not establish a benefit from longer interaction, and response length does not measure hidden reasoning.

The systems also organize inspection and computation differently. Codex can combine these operations, whereas Claude Code exposes separate operations for reading, editing, and execution. Available records do not establish whether these differences explain the scores; that would require matched trajectories from both systems. Appendix~\ref{newapp:harness-process} gives the recovery example and counting definitions.

\begin{table}[!htbp]
\caption{Cross-harness performance on 250 tasks.}
\label{tab:ablation-evidence-harness}
\centering
\small
\setlength{\tabcolsep}{5pt}
\renewcommand{\arraystretch}{1.08}
\begin{tabular*}{\linewidth}{@{\extracolsep{\fill}}llrrr@{}}
\toprule
Model & Access & Codex & Claude Code & $\Delta$ (pp) \\
\midrule
GPT-6 Astra & Harness + 3D & 0.5291 & 0.4082 & $+12.09$ \\
 & Harness + 3D + MCP & 0.5282 & 0.3942 & $+13.40$ \\
\addlinespace[2pt]
GPT-6.1 Sol & Harness + 3D & 0.4893 & 0.3865 & $+10.28$ \\
 & Harness + 3D + MCP & 0.5110 & 0.3843 & $+12.67$ \\
\addlinespace[2pt]
Claude Fable 5.1 & Harness + 3D & 0.5075 & 0.4619 & $+4.56$ \\
 & Harness + 3D + MCP & 0.5349 & 0.4631 & $+7.18$ \\
\addlinespace[2pt]
Claude Opus 5.5 & Harness + 3D & 0.5448 & 0.4539 & $+9.10$ \\
 & Harness + 3D + MCP & 0.5322 & 0.4343 & $+9.78$ \\
\bottomrule
\end{tabular*}
\par\smallskip
\begin{minipage}{\linewidth}\footnotesize
$\Delta$ denotes Codex $-$ Claude Code in percentage points.
\end{minipage}
\end{table}

\subsection{Fixed-view budget and camera plan}
\label{newsec:camera-ablation}
In the earlier camera study, more views did not consistently improve scores. Aligned views changed little from three to six ($0.1516$ to $0.1529$) and decreased at twelve; orbit views tied at six and twelve (Table~\ref{tab:ablation-camera-core}). Six views provide a practical comparison point for that study, but are not an established optimum for the present benchmark.

The study tested GPT-6 Astra with Claude Code and LiteLLM on 250 tasks, using maximum reasoning effort and retaining the last attempt. It compared aligned and orbit views at $N\in\{1,3,6,12\}$. The present evaluation includes only six fixed views. The task set differs, and matching inputs and scoring conditions are not established. We therefore report the earlier pattern separately. Appendix~\ref{newapp:historical} provides the full results and the medical image-count exception.

\begin{table}[H]
\caption{Mean task scores in the earlier camera-plan study using GPT-6 Astra with Claude Code and LiteLLM.}
\label{tab:ablation-camera-core}
\centering
\small
\setlength{\tabcolsep}{6pt}
\renewcommand{\arraystretch}{1.08}
\begin{tabular*}{\linewidth}{@{\extracolsep{\fill}}lrrrr@{}}
\toprule
Camera plan & $N=1$ & $N=3$ & $N=6$ & $N=12$ \\
\midrule
Aligned & 0.1223 & 0.1516 & 0.1529 & 0.1427 \\
Orbit & 0.1171 & 0.1204 & 0.1463 & 0.1463 \\
\bottomrule
\end{tabular*}
\par\smallskip
\begin{minipage}{\linewidth}\footnotesize
Each cell averages 250 task scores at maximum reasoning effort, retaining the last attempt. $N$ is the nominal number of views; medical inputs contain $2N$ images. The present evaluation includes only six fixed views. Appendix~\ref{newapp:historical} gives full diagnostics and comparison limits.
\end{minipage}
\end{table}

\subsection{Tool-only access versus source computation}
\label{newsec:mcp-only-ablation}
Harness + 3D + MCP permits both source computation and viewer inspection. An MCP-only condition instead restricts evidence acquisition to the inspection interface. The earlier study included this condition, giving a mean score of $0.1757$ for GPT-6 Astra across 250 tasks (Table~\ref{tab:ablation-evidence-mcp}).

That run used Claude Code and LiteLLM at maximum reasoning effort and retained the last attempt. Although it was described as tool-only, the available records cannot verify that file access and direct computation were fully disabled. Its task set and execution conditions also differ from the present evaluation. The score therefore illustrates a distinct access protocol, without estimating the benefit of MCP in Harness + 3D + MCP. Appendix~\ref{newapp:historical} gives the remaining limits.

\begin{table}[H]
\caption{Earlier MCP-only and main 3D harness scores for GPT-6 Astra. The access protocols, task sets, and execution conditions are not matched, so these scores cannot estimate the effect of adding MCP.}
\label{tab:ablation-evidence-mcp}
\centering
\small
\setlength{\tabcolsep}{8pt}
\renewcommand{\arraystretch}{1.10}
\begin{tabular}{@{}llr@{}}
\toprule
Cohort ($N=250$) & Access & Score \\
\midrule
Earlier study & MCP-only & 0.1757 \\
\addlinespace[2pt]
Main evaluation & Harness + 3D & 0.5291 \\
 & Harness + 3D + MCP & 0.5282 \\
\bottomrule
\end{tabular}
\par\smallskip
\begin{minipage}{\linewidth}\footnotesize
Earlier and main results use different cohorts and harnesses.
\end{minipage}
\end{table}

\subsection{Camera-plan comparison}
The camera-plan comparison above gives the main pattern and mean scores. Table~\ref{tab:ablation-evidence-2d} adds per-condition diagnostics and separate present-day Harness + 2D scores. The earlier study contains 250 tasks under eight camera conditions and one MCP-only condition. It shares 205 task identifiers with the present benchmark; the 45 replacement questions are absent.

The present evaluation includes only six fixed views and provides no matched camera-plan comparison. The earlier documentation reports $2N$ medical images per task, half identical between aligned and orbit plans. This limits interpretation of medical camera differences. The original rendering records are unavailable for independent verification.

\begin{table}[H]
\caption{Earlier camera-plan results and present Harness + 2D scores. Panel A compares eight camera conditions; Panel B gives selected Harness + 2D scores from the main evaluation. Task sets, execution conditions, and repeat policies are not matched, so differences between panels cannot estimate a camera effect.}
\label{tab:ablation-evidence-2d}
\centering
\small
\setlength{\tabcolsep}{5pt}
\renewcommand{\arraystretch}{1.08}
\begin{tabular*}{\linewidth}{@{\extracolsep{\fill}}lrrrrrr@{}}
\toprule
\multicolumn{7}{@{}l}{\textit{Panel A: earlier study, GPT-6 Astra with Claude Code and LiteLLM}}\\
Plan & $N$ & Mean & Full & Zero & Ineligible & Retried \\
\midrule
Aligned & 1 & 0.1223 & 24 & 210 & 8 & 10 \\
Aligned & 3 & 0.1516 & 29 & 198 & 8 & 9 \\
Aligned & 6 & 0.1529 & 29 & 198 & 8 & 6 \\
Aligned & 12 & 0.1427 & 26 & 199 & 8 & 2 \\
\midrule
Orbit & 1 & 0.1171 & 23 & 211 & 8 & 3 \\
Orbit & 3 & 0.1204 & 22 & 207 & 8 & 5 \\
Orbit & 6 & 0.1463 & 27 & 199 & 8 & 4 \\
Orbit & 12 & 0.1463 & 28 & 199 & 8 & 2 \\
\bottomrule
\end{tabular*}
\smallskip
\begin{tabular*}{\linewidth}{@{\extracolsep{\fill}}lr@{}}
\toprule
\multicolumn{2}{@{}l}{\textit{Panel B: selected Harness + 2D scores from the main evaluation}}\\
Model & Mean task score \\
\midrule
GPT-6 Astra & 0.2498 \\
GPT-6.1 Sol & 0.2044 \\
Claude Fable 5.1 & 0.2965 \\
Claude Opus 5.5 & 0.2383 \\
\bottomrule
\end{tabular*}
\par\smallskip
\begin{minipage}{\linewidth}\footnotesize
\textbf{Panel A:} Each condition contains 250 tasks, uses maximum reasoning effort, and retains the last attempt. The earlier task set shares 205 identifiers with the present benchmark; the 45 replacement tasks are absent. Questions, references, scoring, and camera inputs are not fully matched. Full and Zero count retained scores of one and zero. Ineligible and Retried count tasks marked ineligible and tasks retried in the original records. These counts can overlap Zero and do not change the 250-task denominator. Medical tasks contain $2N$ images, reportedly with half identical between camera plans; original rendering records are unavailable for verification.
\textbf{Panel B:} The main evaluation covers 12 models and 250 tasks. Eleven models have three scores per task. Hunyuan has two scores for three tasks and three for the other 247. Harness + 2D records do not verify the harness for every run. The main Astra configuration uses Codex, while the earlier study used Claude Code and LiteLLM. Only six views were tested in the main evaluation, so these results do not establish an optimal view budget.
\end{minipage}
\end{table}

\subsection{Tool-only access}
Table~\ref{tab:ablation-evidence-mcp} and the tool-only comparison above describe the earlier MCP-only condition, with a mean of $0.1756667$ across 250 tasks. Its documentation states that agents could inspect raw objects only through the MCP interface. However, available tool restrictions and trajectories are insufficient to verify that direct file access and computation were disabled. This differs from Harness + 3D + MCP, which explicitly permits both source computation and inspection. No matched comparison with the present task set is available.

\subsection{Paired trajectories: when MCP helps and when it does not}
\label{newsec:mcp-trajectories}
The score layer records outcomes, not evidence paths. We therefore inspected paired Harness + 3D and Harness + 3D + MCP trials on the same task in archived runs that retain tool logs. Table~\ref{tab:mcp-paired-cases} summarizes five cases and their call counts. The cases are illustrative and come from earlier or supplementary layers; they do not estimate the main-grid MCP effect.

\begin{table}[H]
\caption{Paired Harness + 3D and Harness + 3D + MCP trials on the same task. Scores are task means over available repeats; the Harness + 3D and Harness + 3D + MCP columns also record the trial-level behavior that produced the score. MCP calls count actual tool invocations recorded by the verifier.}
\label{tab:mcp-paired-cases}
\centering
\footnotesize
\setlength{\tabcolsep}{4pt}
\renewcommand{\arraystretch}{1.12}
\begin{tabularx}{\linewidth}{@{}>{\raggedright\arraybackslash}p{0.19\linewidth} >{\raggedright\arraybackslash}X >{\raggedright\arraybackslash}X >{\raggedright\arraybackslash}p{0.20\linewidth}@{}}
\toprule
Task & Harness + 3D & Harness + 3D + MCP & Reading \\
\midrule
\multicolumn{4}{@{}l}{\textit{GPT-6 Astra with Codex: benchmark scores and archived trials}}\\
Binding geometry (0114) & Mean 0.33; the trial without MCP excluded the metal--histidine pair and reported the next contact at 2.856~\AA{} & Mean 1.00; 10 MCP calls; scene selection and distance measurement returned 2.229~\AA{} & MCP encoded the scoring convention for direct links \\
\addlinespace[2pt]
Conformation (0107) & Mean 1.00; reproduced the Mol* display sign and precision, $+179.84^\circ$ and $0.16^\circ$ & Mean 0.50; 4 MCP calls; copied the raw tool value $-179.84277183911854^\circ$ & Tool returned a number but not the display convention \\
\addlinespace[2pt]
Architecture (exp-02) & Mean 0.50; built a browser-based viewer and returned approximate dimensions & Mean 0.00; 2 MCP calls; scene loading failed for the source format and the viewer was unavailable & Coverage gap; no fallback to source tooling \\
\midrule
\multicolumn{4}{@{}l}{\textit{Claude Fable~5.1, Claude Code, 94-task control}}\\
CAD extents (hard-001-copy) & Replicates 0/1/0; ratio 0.625 written as 0.63, 0.62 and 0.63 & Replicates 1/0/0; all three attempted MCP; ratio 0.625 written as 0.62, 0.63 and 0.63 & Same geometry; the score turns on a rounding boundary \\
\addlinespace[2pt]
Lesion pointing (0028) & Replicates 0/0/0; computed CT slices selected one calcific focus & Replicates 1/1/1; 0 MCP calls; computed CT slices selected another focus & The Harness + 3D + MCP label did not imply MCP use \\
\bottomrule
\end{tabularx}
\par\smallskip
\begin{minipage}{\linewidth}\footnotesize
Astra rows report benchmark means; the binding-geometry behavior comes from an archived Codex trial that reproduces the 2.229~\AA{} reference, and the architecture row is the 20 October rerun. Fable rows report the 94-task Claude Code control at recorded high effort. The cases illustrate mechanisms and are not estimates of the main-grid MCP effect.
\end{minipage}
\end{table}

\paragraph{When MCP helps.}
In the binding-geometry case (0114), the reference contact is $2.229$~\AA{} between the BCL 282 magnesium and His 173 NE2. The archived Harness + 3D trial treated the metal--histidine contact as a directly linked pair, excluded it, and reported the next non-linked contact at $2.856$~\AA{}. The Harness + 3D + MCP trial opened the scene, selected the pair, and used the distance-measurement tool, which returned $2.229$~\AA{}. The MCP inspection also revealed how the measurement was made and that the structure lacked explicit connectivity records. MCP therefore helped by encoding the benchmark's contact convention, not merely by rendering the molecule.

\paragraph{When MCP is used but does not help.}
In the conformation case (0107), the Harness + 3D trial examined how the Mol* molecular viewer calculated and displayed dihedral angles, then returned the reference $+179.84^\circ$ and $0.16^\circ$. The Harness + 3D + MCP trial used the dihedral-angle measurement tool, which returned $-179.84277183911854^\circ$, and copied that raw signed value without converting it to the displayed sign and precision. The signed-angle answer failed the circular tolerance; the deviation answer passed. The tool supplied a number but not the convention that made the number scoreable.

In the architecture case (exp-02), the scene-loading tool could not read the building's source format, and the dedicated viewer was unavailable. Because the question restricted evidence to viewer rotation and screenshots, the Harness + 3D + MCP trial returned no dimensions. The Harness + 3D trial built a browser-based viewer for the scene and returned dimensions close enough for partial credit. This is a coverage and fallback failure, not a failure of visual reasoning.

\paragraph{When the difference is not MCP.}
In the CAD-extents case (hard-001-copy), all six trials in the 94-task control computed the same three extents ($2.00$, $1.84$, $1.25$) and the same ratio $0.625$. The reference requires two decimals; trials that wrote $0.62$ passed, and trials that wrote $0.63$ failed. The Harness + 3D and Harness + 3D + MCP arms each contributed both outcomes, and the arm means were identical ($0.333$). The paired difference is a rounding boundary, not an MCP effect. In the lesion-pointing case (0028), the Harness + 3D + MCP trial never called an MCP tool; both arms generated CT slices computationally, and the two trials selected different calcific foci. The route label did not determine the evidence path.

\paragraph{Prevalence.}
Tool logs are uneven. In the archived 250-task Astra Codex batch, 178 of 250 Harness + 3D + MCP trials called at least one MCP tool (1,878 calls), but the arm mean was $0.3909$ versus $0.3970$ for Harness + 3D. In the 94-task Fable Claude Code control, 84 of 282 Harness + 3D + MCP trials called MCP; the arm means were $0.6344$ versus $0.6310$. In the 3,000-trial Fable/Opus Codex batch, 43\% and 36\% of Harness + 3D + MCP trials called MCP, and the score association was positive for Fable but negative for Opus. These are descriptive comparisons; task selection and trial variance confound them. They are consistent with a mixture of sparse, offsetting, and sometimes non-tool effects, not with a reliable average MCP benefit.

\end{document}